\documentclass{article}

\usepackage[main, final]{neurips_2026}

\usepackage[utf8]{inputenc} % allow utf-8 input
\usepackage[T1]{fontenc}    % use 8-bit T1 fonts
\usepackage{hyperref}       % hyperlinks
\usepackage{url}            % simple URL typesetting
\usepackage{booktabs}       % professional-quality tables
\usepackage{amsfonts}       % blackboard math symbols
\usepackage{nicefrac}       % compact symbols for 1/2, etc.
\usepackage{microtype}      % microtypography
\usepackage{xcolor}         % colors

\newcommand\para[1]{\noindent\textbf{#1}\,}

\usepackage{amsmath}
\usepackage{amssymb}
\usepackage{mathtools}
\usepackage{amsthm}

\usepackage{algorithm}
\usepackage{algpseudocode}
\usepackage{wrapfig}

\usepackage[capitalize,noabbrev]{cleveref}

\theoremstyle{plain}
\newtheorem{theorem}{Theorem}[section]
\newtheorem{proposition}[theorem]{Proposition}
\newtheorem{lemma}[theorem]{Lemma}
\newtheorem{corollary}[theorem]{Corollary}
\theoremstyle{definition}

\theoremstyle{remark}

\crefname{proposition}{Proposition}{Propositions}
\Crefname{proposition}{Proposition}{Propositions}
\crefname{lemma}{Lemma}{Lemmas}
\Crefname{lemma}{Lemma}{Lemmas}
\crefname{corollary}{Corollary}{Corollaries}
\Crefname{corollary}{Corollary}{Corollaries}
\crefname{definition}{Definition}{Definitions}
\Crefname{definition}{Definition}{Definitions}
\crefname{assumption}{Assumption}{Assumptions}
\Crefname{assumption}{Assumption}{Assumptions}
\crefname{remark}{Remark}{Remarks}
\Crefname{remark}{Remark}{Remarks}

\title{Rethinking Knowledge Distillation for Diffusion Language Models}

\author{%
  \textbf{%
    Yuxuan Sun\textsuperscript{1} \quad
    Yuanjian Xu\textsuperscript{2} \quad
    Jianing Hao\textsuperscript{2} \quad
    Yu Li\textsuperscript{3}}\\
  \textbf{%
    Sowmen Das\textsuperscript{4} \quad
    Zhong Li\textsuperscript{5} \quad
    Sangwoong Yoon\textsuperscript{6}\thanks{Co-senior authors.} \quad
    Miguel R. D. Rodrigues\textsuperscript{1}\footnotemark[1]}\\[0.5em]
  \textsuperscript{1}University College London \quad
  \textsuperscript{2}HKUST \quad
  \textsuperscript{3}Peking University\\
  \textsuperscript{4}MediaTek Research UK \quad
  \textsuperscript{5}MSRA \quad
  \textsuperscript{6}UNIST\\
  \vspace{-0.5em} \\
  \texttt{ys507@cantab.ac.uk}
}

\begin{document}

\maketitle

\begin{abstract}
Knowledge distillation has recently proven highly effective for improving autoregressive models, but its applicability to DLMs remains poorly understood. We present the first systematic study of KD for DLMs and uncover two non-intuitive failures of classical forward-KL (FKL) distillation: (i) student perplexity does not improve monotonically with more teacher-generated data, and (ii) on the same teacher-generated corpus, FKL distillation underperforms simply training the student with the original DLM denoising objective. 
To address them, we propose a unified, data-free distillation framework for masked DLMs. First, the framework runs an off-policy distillation phase that absorbs general knowledge from a static set of teacher-generated data. Then, the framework employs on-policy distillation which consists of three techniques: a trajectory loss along the student's own reverse process, confidence reweighting that down-weights tokens where the teacher is unsure, and a trust-region anchor that stabilizes the moving target. 
On MDLM, the resulting student attains a GPT-2 perplexity of $24.75$, surpassing the teacher's $27.04$; on Dream-7B distillation to a 93M student model, the framework reduces Qwen-2.5 perplexity from the FKL-distilled baseline of $25.23$ to $19.83$ while preserving performance on zero-shot general capability benchmarks.
\end{abstract}

\section{Introduction}
\label{sec:intro}

Discrete diffusion language models (DLMs) have recently emerged as a compelling alternative to autoregressive large language models, supporting parallel generation while remaining competitive in modeling quality \citep{sahoo2024simple,lou2023discrete,nie2025large,ye2025dream}. As frontier DLMs grow larger, deploying them in latency- or memory-constrained settings, such as on-device assistants, edge systems, and robotic systems with limited accelerators, increasingly motivates compressing them into much smaller students. Knowledge distillation (KD) \citep{hinton2015distilling} is the canonical tool for this compression. KD has also been applied in modern AR LLM post-training, where distilling from a strong teacher often yields larger gains in reasoning ability than reinforcement learning \citep{yue2025does,yang2025qwen3,deepseekai2026deepseekv4}. This raises a natural question:
can knowledge distillation be made similarly effective for closing the teacher--student gap in the DLM paradigm? \looseness=-1

We study this question in the data-free setting, where the student is trained solely on samples synthesized by the teacher rather than on tokens from a real dataset. Two considerations motivate this choice. First, the training corpora of frontier DLMs are typically inaccessible \citep{song2025seed, labs2025mercury}, making it important to distill without access to real corpora. Second, a static real-text corpus can give an incomplete or misaligned picture of what the teacher actually generates, since the teacher often generalizes beyond its original training set or has been shifted away from it by post-hoc finetuning. However, teacher-sampled trajectories can faithfully expose the teacher's generative dynamics, which is what the student should match \citep{tong2025flow}.

\begin{wrapfigure}{r}{0.47\textwidth}
    \vspace{-8pt}
    \centering
    \setlength{\abovecaptionskip}{2pt}   % default is 10pt                              
    \setlength{\belowcaptionskip}{0pt}  
    \includegraphics[width=\linewidth]{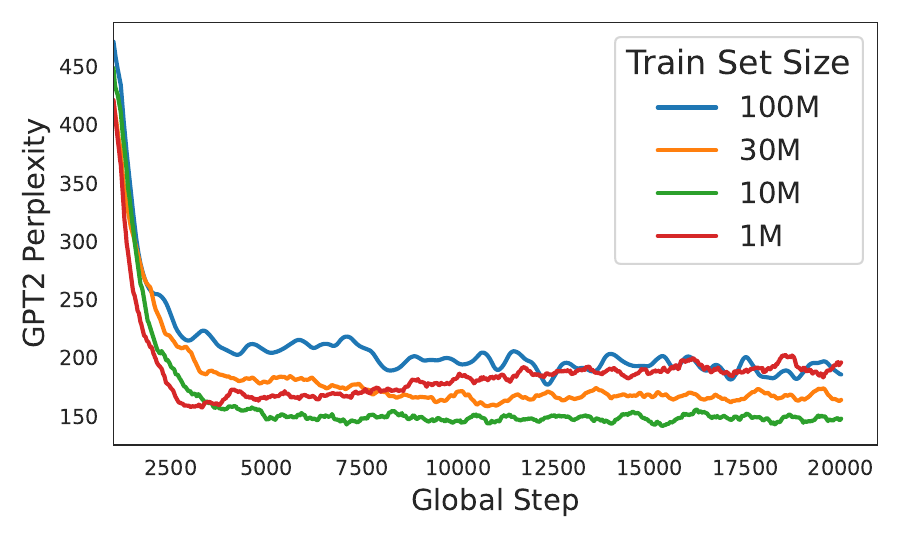}
    \caption{\textbf{Scaling failure of forward-KL distillation in MDLM.} GPT-2 perplexity of an MDLM student does not improve monotonically with the size of the teacher-generated corpus: $10$M tokens are optimal, while larger sets ($30$M, $100$M) yield worse perplexity. Dream exhibits the same pattern (\Cref{tab:dream_scaling}).}
    \label{fig:forwardkl_fail_smoothed}
    \vspace{-10pt}
\end{wrapfigure}

We present the first systematic study of KD for diffusion language models. Applying classical Forward Kullback--Leibler (FKL) distillation, the objective used in the original KD paper \citep{hinton2015distilling}, to MDLM \citep{sahoo2024simple} and Dream \citep{ye2025dream} reveals two phenomena that contradict standard KD intuition:
\begin{itemize}
  \item \textbf{Non-monotone scaling.} Student perplexity does not improve monotonically with more teacher-generated data. For MDLM, $10$M tokens are best, while $30$M and $100$M are worse, as shown in \Cref{fig:forwardkl_fail_smoothed}. Dream exhibits the same non-monotone pattern, shown in \Cref{tab:dream_scaling}.
  \item \textbf{Objective mismatch.} Trained on the same teacher-generated corpus, FKL distillation underperforms simply training the student with the original DLM denoising objective. The same gap is observed for both MDLM and Dream.
\end{itemize}

We trace both failures to the same root cause: FKL is applied only at states drawn from the forward noising process of teacher samples, whereas during inference the student generates samples through its own reverse process. As more teacher-generated tokens are added, the student fits the forward-noised teacher states ever more tightly, while its sampling-time states drift away from them; the off-policy loss provides no corrective signal in this drifted region. This is precisely the ``no implicit prior'' issue raised by \citet{sahoo2025diffusion}: in masked DLMs the noise-to-data map is not deterministic, so the states the student visits at sampling time cannot be derived simply by forward-noising teacher samples. The phenomenon also mirrors recent explanations of hallucination in AR LLMs \citep{kalai2025language}, where objectives that uniformly reward coverage of the teacher’s full distribution also implicitly reward guessing in regions that the student never visits at inference time.

Building on this diagnosis, we propose a unified, data-free distillation framework for masked DLMs that explicitly aligns training-time states with the student's reverse-process occupancy. The framework has two phases. An \emph{off-policy} phase absorbs general teacher knowledge from a static, teacher-generated corpus. Once it stops improving on a held-out evaluator, training switches to an \emph{on-policy} phase whose loss combines three ingredients: (i)~a trajectory loss that supervises the student at multiple intermediate states sampled from its own reverse process, breaking the single-step, single-state bottleneck of FKL; (ii)~confidence reweighting that scales each token's loss by the teacher's max-probability, focusing learning on states where the teacher is reliable; and (iii)~a trust-region anchor on a periodically refreshed copy of the student that stabilizes optimization against the moving rollout distribution. For the smaller MDLM teacher, where FKL is independently too weak, we further replace FKL with the $\alpha$-$\beta$ divergence of \citet{wang2025abkd} in both phases; for the much stronger Dream-7B teacher, FKL suffices.

\textbf{Contributions.}
\begin{enumerate}
  \item The first systematic study of KD for DLMs across two representative scale regimes, identifying two failure phenomena of FKL distillation for both a 110M (MDLM) and a 7B (Dream) teacher model (\Cref{sec:challenges}).
  \item A unified, data-free distillation framework combining off- and on-policy phases, a trajectory loss, confidence reweighting, and a trust-region anchor, with explicit choices of divergence ($\alpha$-$\beta$ divergence for MDLM, FKL for Dream) motivated by teacher capacity (\Cref{sec:method}).
  \item Empirical results: on MDLM, the distilled $10.5$M student reaches GPT-2 perplexity $24.75$, below the $27.04$ of the $110$M teacher; on Dream-7B distillation into a 93M student, the framework reduces Qwen-2.5 perplexity from $25.23$ (FKL) to $19.83$ while preserving zero-shot benchmark accuracy. Ablations isolate the contribution of each component (\Cref{sec:experiments}).
\end{enumerate}

\section{Background and Related Work}
\label{sec:background}

\para{Discrete diffusion language models.}
A masked discrete DLM defines a forward process that progressively replaces tokens of a clean sequence $x_0$ with a special $\mathtt{[MASK]}$ token according to a schedule indexed by $t\in[0,T]$, and learns a parameterized denoiser $f_\theta(x_t,t)$ that predicts the clean tokens at masked positions. Recent work has advanced DLMs in several directions. \citet{lou2023discrete} introduced a discrete diffusion model trained with a score entropy objective derived from Bregman divergence. \citet{sahoo2024simple} proposed MDLM, a masked diffusion model optimized with a Rao-Blackwellized objective, establishing a new state of the art in DLM performance. LLaDA \citep{nie2025large} and Dream \citep{ye2025dream} have scaled masked DLMs to 8B and 7B parameters respectively, demonstrating performance competitive with autoregressive models of similar scale.

\para{Knowledge distillation.}
Classical KD \citep{hinton2015distilling} matches a student's output distribution to that of a fixed teacher under a forward KL divergence. \citet{wang2025abkd} introduce $\alpha$-$\beta$ divergence, which interpolates between FKL and reverse KL via two scalars controlling hardness concentration and confidence concentration; we adopt ABKL on MDLM where FKL underperforms. In the AR model setting, several approaches have refined distillation objectives. \citet{gu2023minillm} introduced MiniLLM, which minimizes reverse KL divergence between teacher and student distributions. \citet{ko2024distillm} proposed DISTILLM, employing a skew KL divergence and improving sample efficiency through a replay buffer mechanism; \citet{lu2025onpolicydistillation} highlight on-policy distillation, where the student is trained on data drawn from itself rather than from a static corpus. \citet{agarwal2024policy} explore curriculum mixing of off- and on-policy losses for AR models; we revisit this in our ablations.

\para{Distillation for DLMs.}
Research on distilling diffusion language models remains limited and has primarily focused on reducing the number of inference sampling steps. For instance, SDTT \citep{deschenaux2025beyond} aligns student and teacher log-probabilities to reduce sampling steps from 1024 to 64. DI4C \citep{hayakawa2025distillation} models dimensional correlations to distill to one or a few steps. Our setting is orthogonal: we keep the sampling schedule and instead distill a much smaller student. This complements step distillation and could in principle be combined with it.

\section{Two Failures of Forward-KL Distillation for DLMs}
\label{sec:challenges}

We use pretrained state-of-the-art DLMs as teachers and distill them into smaller students: an MDLM teacher (110M) and a Dream teacher (7B). For each teacher, we generate a static corpus by unconditional sampling and train a small student to minimize the forward KL between teacher and student output distributions on these samples. The MDLM student has 10.5M parameters and follows teacher's DiT architecture \citep{peebles2023scalable}; the Dream student has 93M parameters and follows the Dream architecture (\Cref{tab:student_architectures}).
To rule out distributional shift as a confounder, we use nested data subsets. The MDLM corpora satisfy $\mathcal{D}_{1\mathrm{M}}\subset\mathcal{D}_{10\mathrm{M}}\subset\mathcal{D}_{30\mathrm{M}}\subset\mathcal{D}_{100\mathrm{M}}$, and the Dream corpora satisfy $\mathcal{D}_{30\mathrm{M}}\subset\mathcal{D}_{100\mathrm{M}}\subset\mathcal{D}_{300\mathrm{M}}\subset\mathcal{D}_{500\mathrm{M}}\subset\mathcal{D}_{1000\mathrm{M}}$, so any non-monotone scaling cannot be attributed to distributional shift across scales.

\textbf{Finding 1: Non-monotone scaling}
For MDLM, FKL-distilled students attain GPT-2 perplexities at $\{1\mathrm{M}, 10\mathrm{M}, 30\mathrm{M}, 100\mathrm{M}\}$ shown in \Cref{fig:forwardkl_fail_smoothed}: $10$M is optimal, and additional data harms perplexity. The same non-monotonicity is observed when distilling Dream into a 93M student (\Cref{tab:dream_scaling}): the $300$M-token corpus is optimal at $25.23$, while the $500$M and $1000$M corpora regress.

\textbf{Finding 2: Objective mismatch}
Holding the corpus fixed, FKL distillation underperforms simply training the student with the teacher's own denoising objective on the same data. For MDLM at the optimal $10$M corpus, FKL distillation yields $122.37$ versus $90.20$ for the original MDLM objective (\Cref{tab:forwardkl_realdata}). The same pattern holds for Dream at the optimal $300$M corpus: FKL distillation reaches $25.23$ while original Dream denoising objective on the same teacher-generated tokens reaches $20.71$.

This diagnosis is not merely empirical: we later show in \Cref{prop:offpolicy_fail} that a student can attain zero FKL at every forward-visited state yet have total-variation distance arbitrarily close to $1$ from the data, because its reverse process visits states the forward process never reaches. This motivates the method in \Cref{sec:method}: we replace off-policy supervision on forward-noised teacher samples with on-policy supervision along the student's own trajectory, weighting the teacher signal by confidence and stabilizing updates with a trust region.

\begin{table}[t]
  \centering
  \begin{minipage}[t]{0.52\textwidth}
    \centering
    \caption{Dream-7B FKL distillation to a 93M student exhibits non-monotone corpus scaling. Lower is better.}
    \label{tab:dream_scaling}
    \begin{footnotesize}
    \setlength{\tabcolsep}{4pt}
    \begin{tabular}{@{}lccccc@{}}
      \toprule
      Corpus size & 30M & 100M & 300M & 500M & 1000M \\
      \midrule
      Qwen-2.5 PPL $\downarrow$ & 46.58 & 33.54 & \textbf{25.23} & 44.48 & 37.68 \\
      \bottomrule
    \end{tabular}
    \end{footnotesize}
  \end{minipage}\hfill
  \begin{minipage}[t]{0.46\textwidth}
    \centering
    \caption{Objective mismatch for MDLM at the $10$M-token corpus. Lower is better.}
    \label{tab:forwardkl_realdata}
    \begin{footnotesize}
    \setlength{\tabcolsep}{4pt}
    \begin{tabular}{@{}lr@{}}
      \toprule
      Method & GPT-2 PPL $\downarrow$ \\
      \midrule
      FKL distillation on real OWT & 138.84 \\
      FKL distillation on teacher samples & 122.37 \\
      MDLM objective on teacher samples & \textbf{90.20} \\
      \bottomrule
    \end{tabular}
    \end{footnotesize}
  \end{minipage}
\end{table}

\section{Method}
\label{sec:method}

\subsection{Two-phase pipeline}
Let $q_t(\cdot\mid x_0)$ denote the masked-diffusion forward kernel that, given a clean sequence $x_0$ and time $t\!\in\![0,T]$, produces a partially masked sequence $x_t$. Let $p_T(\cdot\mid x_t,t)$ and $p_\theta(\cdot\mid x_t,t)$ denote the teacher's and student's per-position output distributions at time $t$ given $x_t$, and let $D(\cdot\,\Vert\,\cdot)$ be a divergence; in our experiments $D$ is FKL for Dream and $\alpha$-$\beta$ divergence for MDLM.

We use a two-phase pipeline. The first phase, \emph{off-policy distillation}, draws a clean sequence $x_0$ from a static teacher-generated corpus $\mathcal{D}_{\text{off}}$, samples a time $t$ and a noised state $x_t \sim q_t(\cdot\mid x_0)$, and matches the teacher and student distributions at $x_t$:
\begin{equation}
\mathcal{L}_{\text{off}}(\theta) \;=\; \mathbb{E}_{x_0\sim\mathcal{D}_{\text{off}},\; t\sim\mathcal{U}(0,T),\; x_t\sim q_t(\cdot\mid x_0)}\!\left[\,D\!\left(p_T(\cdot\mid x_t,t)\,\Vert\,p_\theta(\cdot\mid x_t,t)\right)\right].
\label{eq:loff}
\end{equation}

\begin{wrapfigure}{r}{0.46\textwidth}
    \vspace{-8pt}
    \centering
    \setlength{\abovecaptionskip}{2pt}   % default is 10pt
    \setlength{\belowcaptionskip}{0pt}  
    \includegraphics[width=\linewidth]{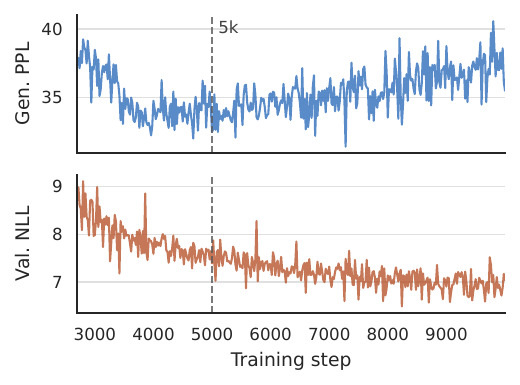}
    \vspace{-1em}
    \caption{\textbf{Off-policy $\alpha$-$\beta$ divergence overfitting dynamics in MDLM.} Validation NLL on OWT keeps decreasing, but GPT-2 generative perplexity rises after the 5k-step checkpoint.}
    \label{fig:abkl_overfit_dynamics}
    \vspace{-1pt}
\end{wrapfigure}
We use a static corpus rather than on-the-fly teacher generation because
(i) teacher generation is the expensive step, and decoupling it from training improves throughput, and (ii) a static corpus extracts more signal per generated sample than discarding it after a single optimizer step.

We empirically observe that prolonged off-policy training overfits: the student's negative log-likelihood on OWT (the teacher's training data) keeps decreasing while its perplexity on generic text begins to rise. \Cref{fig:abkl_overfit_dynamics} shows this behavior for MDLM with $\alpha$-$\beta$ divergence: after the 5k-step checkpoint, validation NLL continues to decrease while GPT-2 generative perplexity rises. This is the symptom of the off-policy/on-policy mismatch diagnosed in \Cref{sec:challenges}. Once this divergence appears, we switch to the \emph{on-policy} phase, initialized from the off-policy checkpoint.

\begin{figure}[t]
  \centering
  \includegraphics[width=0.99\columnwidth]{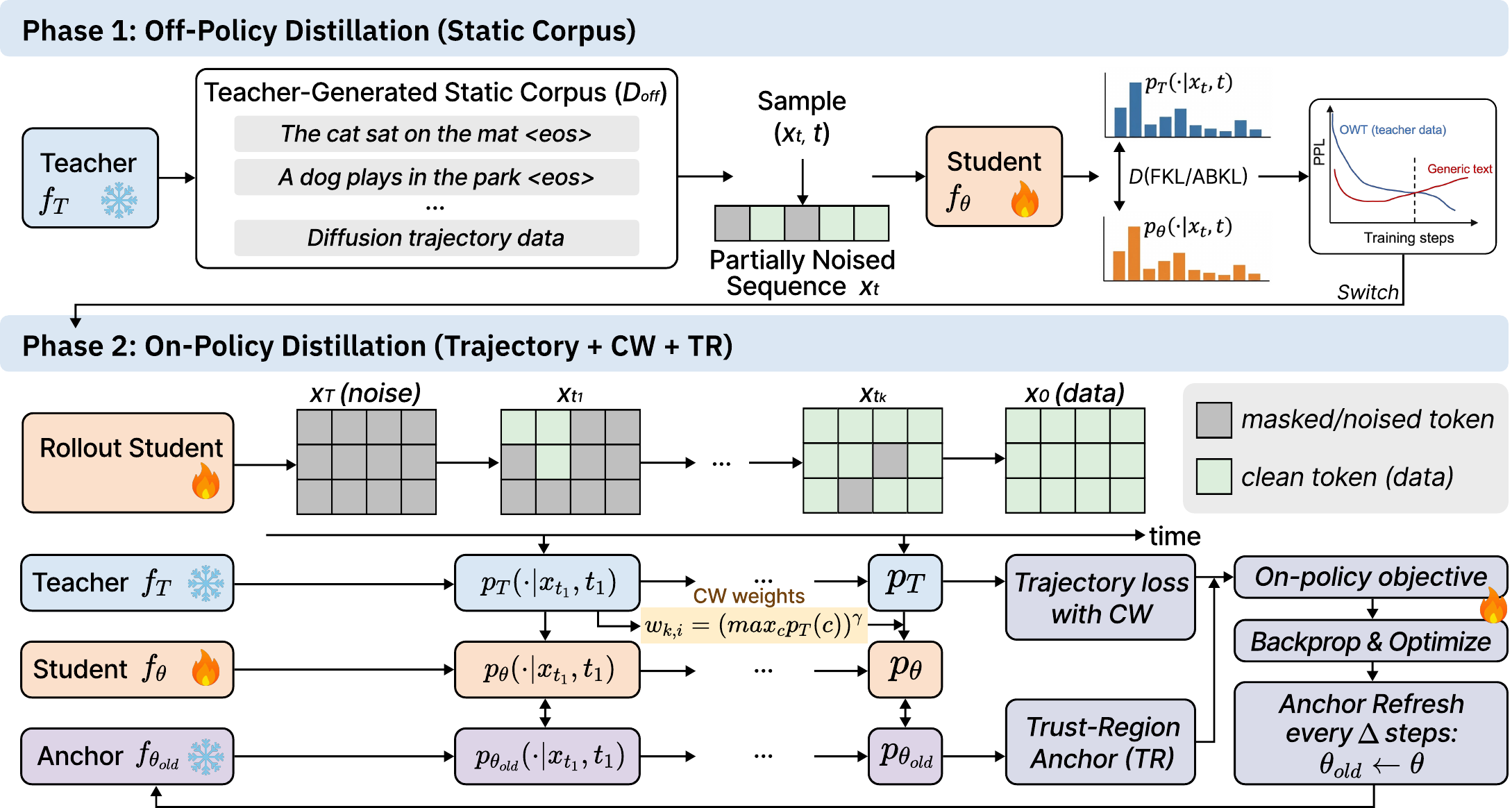}
  \caption{\textbf{Two-phase distillation pipeline.} Phase 1 (off-policy): the student is trained on a static teacher-generated corpus with divergence $D$. When held-out perplexity stops improving, training switches to Phase 2 (on-policy): the student rolls out partial reverse trajectories, intermediate states $x_{t_k}$ are scored by both teacher and student under a trajectory loss, the loss is reweighted by teacher confidence (CW), and a trust-region KL anchor (TR) against a periodically refreshed copy $\theta_{\text{old}}$ stabilizes optimization.}
  \label{fig:pipeline}
\end{figure}

The on-policy loss combines confidence-weighted trajectory matching with a trust-region anchor:
\begin{equation}
\boxed{\;
\mathcal{L}_{\text{on}}(\theta) \;=\; \mathcal{L}_{\text{traj}}^{\text{CW}}(\theta) \;+\; \lambda\;\mathbb{E}_{x_t\sim\rho_{\theta}^t}\!\left[\,\mathrm{KL}\!\left(p_{\theta_{\text{old}}}(\cdot\mid x_t,t)\,\Vert\,p_\theta(\cdot\mid x_t,t)\right)\right]
\;}
\label{eq:lon}
\end{equation}
where $\mathcal{L}_{\text{traj}}^{\text{CW}}$ is the confidence-weighted trajectory loss defined in \Cref{sec:traj,sec:cw}, and the second term is trust-region anchor of \Cref{sec:tr}. The full pipeline is shown in \Cref{fig:pipeline} and Algorithm~\ref{alg:onpolicy}.

\subsection{Trajectory loss along the student's reverse process}
\label{sec:traj}
On-policy training in our setting works as follows: the student rolls out a partial reverse trajectory with the same number of diffusion steps used at inference; from this trajectory, we sample $K$ intermediate states $x_{t_1},\dots,x_{t_K}$ by drawing the timesteps $t_1,\dots,t_K$ uniformly at random; both teacher and student are queried on each $x_{t_k}$. The trajectory loss is
\begin{equation}
\mathcal{L}_{\text{traj}}(\theta) \;=\; \mathbb{E}_{x_0\sim\rho_\theta,\;\{t_k\}\sim\mathcal{U}\{1..T\}^{\uparrow}}\!\left[\,\frac{1}{K}\sum_{k=1}^{K} D\!\left(p_T(\cdot\mid x_{t_k},t_k)\,\Vert\,p_\theta(\cdot\mid x_{t_k},t_k)\right)\right].
\label{eq:ltraj}
\end{equation}
The trajectory states $\{x_{t_k}\}$ are treated as stop-gradient: they are produced inside a \texttt{torch.no\_grad()} block during rollout, and only the divergence is differentiated.

\para{Why trajectory matters.}
A single-timestep loss provides only local supervision and gives no signal at the intermediate states that the student actually visits during sampling. Without trajectory supervision, the student can develop self-reinforcing errors: a mistake at one denoising step corrupts the input to the next, and a single random timestep is unlikely to surface that mistake. Supervising multiple intermediate states constrains the student to remain consistent with the teacher across the reverse process, reducing error accumulation.

\para{Bias--variance trade-off in $K$.}
Empirically, $K=4$ performs best: $K=1$ gives a noisy estimate of the trajectory loss, whereas $K\geq 8$ tends to over-constrain the update (\Cref{tab:onpolicy_trajactory}). The reason is that increasing $K$ improves trajectory coverage but also introduces many correlated training signals from the same reverse trajectory. Writing the gradient variance as $\mathrm{Var}\!\left[\nabla\mathcal{L}_{\text{traj}}\right] = \frac{1}{K^2}\sum_{i,j}\mathrm{Cov}\!\left[\nabla\ell_{t_i},\nabla\ell_{t_j}\right]$. There are only $K$ diagonal variance terms but $K(K-1)$ off-diagonal covariance terms. Since all $K$ intermediate timesteps are sampled from the same student trajectory, their gradients are positively correlated; therefore, the off-diagonal terms do not vanish. Consequently, the variance saturates near the average pairwise covariance rather than decreasing at the ideal $1/K$ Monte Carlo rate.

\subsection{Confidence reweighting}
\label{sec:cw}
On-policy rollouts can visit states $x_t$ that lie far from the teacher's training-time states. At such states, the teacher distribution $p_T$ is uncertain, so enforcing the divergence uniformly can inject high-variance and potentially biased supervision. We therefore use token-level confidence reweighting:
\begin{equation}
\mathcal{L}_{\text{traj}}^{\text{CW}}(\theta) \;=\; \mathbb{E}\!\left[\,\frac{1}{K}\sum_{k=1}^{K}\sum_{i} w_{k,i}\,D\!\left(p_{T,k,i}\,\Vert\,p_{\theta,k,i}\right)\right], \quad w_{k,i} \;=\; \big(\max_{c} p_{T,k,i}(c)\big)^{\gamma},
\label{eq:cw}
\end{equation}
where $i$ indexes token positions and $\gamma \geq 0$ controls the strength of reweighting. We find $\gamma=1$ works best in our experiments.
This weighting down-weights supervision at states where the teacher is uncertain. In particular, when the teacher distribution is close to uniform over the vocabulary, $\max_c p_T(c)$ is close to $1/C$, so $w_{k,i}$ becomes small. Confidence reweighting therefore reduces the influence of unreliable teacher targets while preserving stronger supervision at states where the teacher assigns high probability to a small set of tokens.

\subsection{Trust-region anchor}
\label{sec:tr}
Because on-policy training has a moving data distribution, the student can drift away from a working solution, especially when CW down-weights large fractions of a batch. We add a forward-KL anchor to a periodically refreshed copy $\theta_{\text{old}}$ of the student. The full on-policy objective is \cref{eq:lon}, repeated here:
\begin{equation}
\mathcal{L}_{\text{on}}(\theta) \;=\; \mathcal{L}_{\text{traj}}^{\text{CW}}(\theta) \;+\; \lambda\,\mathbb{E}_{x_t\sim\rho_{\theta}^t}\!\left[\,\mathrm{KL}\!\left(p_{\theta_{\text{old}}}(\cdot\mid x_t,t)\,\Vert\,p_\theta(\cdot\mid x_t,t)\right)\right].
\end{equation}
The trust-region states are the same on-policy rollout states $x_t\sim\rho_\theta$ used by the trajectory loss; only the anchor distribution inside the KL is computed under the frozen $\theta_{\text{old}}$. We use a constant $\lambda$ and refresh $\theta_{\text{old}} \!\leftarrow\! \theta$ every $\Delta$ optimizer steps. We only tested FKL for the anchor: the goal here is a soft regularizer that keeps the student close to a recently known-good policy. Schedules for $\lambda$, alternative anchor divergences, and adaptive $\Delta$ are interesting directions we leave to future work.

\begin{algorithm}[t]
\caption{Two-phase data-free distillation for masked DLMs.}
\label{alg:onpolicy}
\begin{algorithmic}[1]
\Require teacher $f_T$, student $f_\theta$, static teacher corpus $\mathcal{D}_{\text{off}}$, divergence $D$, hyperparameters $K,\gamma,\lambda,\Delta$, switch criterion $\mathcal{C}$.
\State \textbf{Off-policy phase.} Train $\theta$ on $\mathcal{D}_{\text{off}}$ by minimizing $\mathcal{L}_{\text{off}}(\theta)$ in \Cref{eq:loff} until $\mathcal{C}$ indicates that held-out evaluator perplexity no longer improves.
\State Initialize $\theta_{\text{old}} \leftarrow \theta$ and refresh counter $c \leftarrow 0$.
\While{on-policy training has not converged}
  \State Roll out the current student reverse process from noise; sample and sort $K$ intermediate states $\{(x_{t_k},t_k)\}_{k=1}^K$. \Comment{stop-gradient on states}
  \State Query the teacher and student to obtain $p_T(\cdot\mid x_{t_k},t_k)$ and $p_\theta(\cdot\mid x_{t_k},t_k)$ for all $k$.
  \State Set $w_{k,i} \leftarrow (\max_c p_{T,k,i}(c))^\gamma$ and compute $\mathcal{L}_{\text{traj}}^{\text{CW}} \leftarrow \tfrac{1}{K}\sum_k\sum_i w_{k,i}\,D(p_{T,k,i}\Vert p_{\theta,k,i})$.
  \State Query the anchor model $p_{\theta_{\text{old}}}(\cdot\mid x_{t_k},t_k)$ and compute $\mathcal{L}_{\text{TR}} \leftarrow \tfrac{1}{K}\sum_k\mathrm{KL}(p_{\theta_{\text{old}},k}\Vert p_{\theta,k})$.
  \State Update $\theta$ using $\nabla_\theta(\mathcal{L}_{\text{traj}}^{\text{CW}}+\lambda\mathcal{L}_{\text{TR}})$.
  \State $c \leftarrow c+1$; if $c\bmod\Delta=0$, refresh $\theta_{\text{old}}\leftarrow\theta$.
\EndWhile
\end{algorithmic}
\end{algorithm}

\subsection{Theoretical analysis}
\label{sec:theory}

We next make the preceding intuition precise. The analysis shows how off-policy distillation incurs a mismatch between training and sampling states, and how on-policy supervision, confidence reweighting, and the trust-region anchor mitigate this mismatch. Full proofs are deferred to \Cref{app:theory}.

\begin{proposition}[Off-policy failure for masked DLMs]
\label[proposition]{prop:offpolicy_fail}
For any $\epsilon>0$, there exists a data distribution over sequences of length $n$, a masked diffusion process with $T\!=\!2$ steps, and a student that achieves zero off-policy divergence at every forward-visited state, yet whose generation quality satisfies $\mathrm{TV}(p_\theta^{\mathrm{gen}},\, p_{\mathrm{data}}) > 1 - \epsilon$.
\end{proposition}

\noindent The proof in \Cref{app:offpolicy_fail} uses a two-token vocabulary where forward noising preserves token correlations, but the student's reverse process breaks these correlations by sampling positions independently, creating states outside forward support. This formalizes the non-monotone scaling of \Cref{sec:challenges}.

\para{Unified bound for on-policy distillation.}
Our main result shows that on-policy trajectory loss with confidence reweighting yields a controlled bound on the generation quality gap; the trust-region anchor separately stabilizes the moving rollout distribution during optimization.

\begin{theorem}[Calibrated divergence theorem]
\label{thm:main}
Let $\ell:\mathcal{V}^L\!\to [0,B]$ be any bounded evaluation function. Let $D$ be a divergence satisfying the calibration condition
\begin{equation}
\mathrm{TV}(P,Q) \;\leq\; \kappa_D\sqrt{D(P\Vert Q)}
\label{eq:calibrated-divergence}
\end{equation}
for all teacher/student one-step conditionals encountered under the student rollout distribution. Define the confidence weight $w_t(x_t) = \bigl(\max_c p_T(c\mid x_t,t)\bigr)^{\gamma}$, and assume the \emph{teacher-calibration condition}: there exists $\varepsilon_0\geq 0$ such that
\begin{equation}
\varepsilon_{\mathrm{lowconf}} \;\coloneqq\; \frac{1}{T}\sum_{t=1}^{T}\mathbb{E}_{x_t\sim q_\theta}\!\bigl[(1-w_t(x_t))\,D_t(x_t)\bigr] \;\leq\; \varepsilon_0,
\label{eq:lowconf-bound}
\end{equation}
where $D_t(x_t)\coloneqq D(p_T(\cdot\mid x_t,t)\Vert p_\theta(\cdot\mid x_t,t))$. Then for any student $\theta$:
\begin{equation}
\underbrace{\bigl|\mathbb{E}_{q_\theta}[\ell(x_0)] - \mathbb{E}_{p}[\ell(x_0)]\bigr|}_{\text{generation gap}}
\;\leq\; BT\kappa_D\;\sqrt{\,
\underbrace{\mathcal{L}_{\mathrm{traj}}^{\mathrm{CW}}}_{\textbf{(I)}\;\text{trajectory}}
+ \underbrace{\varepsilon_0}_{\textbf{(II)}\;\text{low-confidence residual}}
\,},
\label{eq:main-bound}
\end{equation}
where $\mathcal{L}_{\mathrm{traj}}^{\mathrm{CW}}$ is the time-averaged CW trajectory loss under divergence $D$.
\end{theorem}

\noindent Term~(I) is minimized directly by the trajectory loss under the student's own states. Term~(II) is the low-confidence residual: the divergence accumulated on positions the teacher itself flags as uncertain ($w_t\to 0$). Assumption \eqref{eq:lowconf-bound} is a \emph{teacher-strength condition}: a sufficiently capable teacher whose confidence is calibrated with its reliability rarely places non-trivial mass on uncertain positions in regions the student visits, so $(1-w_t)\,D_t$ is uniformly small and \eqref{eq:lowconf-bound} holds with a small $\varepsilon_0$. Empirically this is the regime in which our bound is tight. The trust-region penalty $\mathcal{L}_{\mathrm{TR}}$ does not appear in this bound; its role is to stabilize optimization by controlling the non-stationarity of the rollout distribution between refreshes (analogous to TRPO~\citep{schulman2015trust}; see \Cref{app:tr_convergence}). 

For Dream, FKL satisfies \Cref{eq:calibrated-divergence} with $\kappa_D=1/\sqrt{2}$ by Pinsker's inequality. For MDLM, we use $\alpha$-$\beta$ divergence with $(\alpha,\beta)=(0.2,0.5)$, hence $\alpha+\beta=0.7\leq 1$, so \Cref{lem:abkl-pinsker} shows that the theorem applies to the $\alpha$-$\beta$ divergence trajectory loss with the same $\kappa_D=1/\sqrt{2}$. 

Our on-policy trajectory loss is an instance of the DAgger framework~\citep{ross2011reduction} adapted to diffusion: the teacher is the ``expert'' queried at the student's visited states. Off-policy distillation corresponds to behavioral cloning and therefore suffers $\mathcal{O} (T^2\epsilon)$ compounding error. In contrast, on-policy training reduces this to $\mathcal{O} (T\epsilon)$, improving the horizon dependence from quadratic to linear. The proof is given in \Cref{app:main_thm}.

\section{Experiments}
\label{sec:experiments}

\subsection{Setup}
We evaluate on two teachers: MDLM-110M \citep{sahoo2024simple} and Dream-7B \citep{ye2025dream}. Both students are smaller versions of their teachers' architectures: a 10.5M-parameter DiT~\citep{peebles2023scalable} for MDLM and a 93M-parameter Dream model for Dream-7B. Full architectures and optimizer hyperparameters are in \Cref{app:implementation}.

For both teachers we run a fully data-free pipeline: we generate samples unconditionally from the teacher, store them as a static corpus $\mathcal{D}_{\text{off}}$, and train the student on this corpus and on its own on-policy rollouts. MDLM samples have length $1024$ (GPT-2 tokenizer); Dream samples have length $256$ (Qwen-2.5 tokenizer). Off-policy uses batch size $512$ (MDLM) and $1024$ (Dream), with learning rates $3\!\times\!10^{-4}$ and $1\!\times\!10^{-4}$ respectively. On-policy is initialized from the converged off-policy checkpoint ($5$k steps for MDLM, $25$k for Dream) and uses smaller batches and a much lower learning rate (batch sizes $16$ and $32$ at $3\!\times\!10^{-8}$, following \citet{lu2025onpolicydistillation}). 

For MDLM we use $\alpha$-$\beta$ divergence with $(\alpha,\beta)=(0.2,0.5)$ throughout; for Dream we use FKL. We adopt $\alpha$-$\beta$ divergence for MDLM because FKL is itself insufficient there: \Cref{tab:forwardkl_realdata} shows FKL distillation reaching only $122.37$ perplexity, whereas $\alpha$-$\beta$ divergence with $(\alpha,\beta)=(0.2,0.5)$ reaches $32.48$ at the same token budget, as shown in \Cref{tab:offpolicy_abkl}. For Dream, we use FKL because its 7B teacher is substantially stronger than the MDLM teacher. In this setting, the choice of divergence has a smaller effect on performance. We use the same $(\alpha,\beta)$ in both phases for MDLM, having found in experiments that mixing divergences across phases is worse. We use $K\!=\!4$ trajectory samples in both. The anchor is refreshed every $\Delta\!=\!100$ steps (MDLM) and $\Delta\!=\!150$ steps (Dream).

We follow prior work \citep{sahoo2024simple,deschenaux2025beyond,hayakawa2025distillation} and report perplexity under a stronger model with a matched tokenizer. For MDLM, we use GPT-2 Large \citep{radford2019language}. For Dream, we use Qwen-2.5-7B \citep{qwen2.5}, whose tokenizer matches Dream because Dream is initialized from Qwen-2.5.

\subsection{Does teacher-generated data hurt?}
\label{sec:dataquality}
Since FKL distillation underperforms the original denoising objective on the same teacher data (\Cref{sec:challenges}), one hypothesis is that the teacher-generated data is itself worse than real text. We test this directly on MDLM at the optimal $10$M-token budget by training students with FKL on (i) teacher-generated tokens versus (ii) real OWT tokens, and additionally compare to the original MDLM denoising objective on the teacher tokens. Results are shown in \Cref{tab:forwardkl_realdata}: teacher-generated tokens outperform real OWT under FKL distillation. This is consistent with the FreeFlow observation \citep{tong2025flow} that real corpora constrain intermediate diffusion trajectories to the empirical data manifold, while a student should instead follow the teacher's generative dynamics. The remaining gap is therefore not a data problem but a state-distribution problem: FKL is evaluated only on forward-noised teacher samples, whereas at inference time the student must denoise states produced by its own reverse process. Closing this mismatch is precisely what our on-policy framework does.  

\subsection{Main results}
\Cref{tab:main} reports the GPT-2 / Qwen-2.5 perplexity of the distilled student under each ablation of our framework. Off-policy alone is already a strong starting point; switching to on-policy with our trajectory loss at $K\!=\!4$ improves further; adding confidence reweighting and the trust-region anchor gives the best result. On MDLM the final student reaches $24.75$ perplexity, below the $27.04$ of the 110M MDLM teacher. Similar improvements over the teacher have also been observed in continuous diffusion models: SiD~\citep{zhou2024score} reports a distilled student with better FID and Inception Score than its teacher. We hypothesize that the on-policy phase of our pipeline enables the MDLM gain by supervising the student at multiple states along its own reverse trajectory, aligning training more closely with inference than the teacher's pretraining on forward-noised data. On Dream, the final 93M student reaches $19.83$, comfortably below the $25.23$ obtained by FKL distillation alone.

\begin{table}[t]
\caption{Perplexity of the distilled MDLM (10.5M, GPT-2 PPL) and Dream (93M, Qwen-2.5 PPL) students under successive ablations. Lower is better. Best value in each column is in \textbf{bold}.}
\label{tab:main}
\centering
\begin{small}
\begin{tabular}{lcc}
\toprule
Method & MDLM (GPT-2 PPL) $\downarrow$ & Dream (Qwen-2.5 PPL) $\downarrow$ \\
\midrule
Off-policy distillation only & 32.48 & 25.23 \\
\, + on-policy, trajectory $K\!=\!4$ & 25.38 & 24.96 \\
\, + on-policy, traj.\ + CW & 27.44 & 20.88 \\
\, + on-policy, traj.\ + CW + TR & \textbf{24.75} & \textbf{19.83} \\
\midrule
Teacher reference & 27.04 & 12.07 \\
\bottomrule
\end{tabular}
\end{small}
\end{table}

 \begin{wraptable}{r}{0.44\textwidth}
    \vspace{-16pt}
  \caption{MAUVE scores for Dream (93M) students. Higher is better.}
  \label{tab:mauve}
  \centering
  \begin{small}
  \setlength{\tabcolsep}{12pt}
  \begin{tabular}{@{}lc@{}}
  \toprule
  Method & MAUVE $\uparrow$ \\
  \midrule
  Off-policy distillation only & 0.6034 \\
  \, + on-policy, trajectory $K\!=\!4$ & \textbf{0.7643} \\
  \, + on-policy, traj.\ + CW & 0.6787 \\
  \, + on-policy, traj.\ + CW + TR & 0.6677 \\
  \bottomrule
  \end{tabular}
  \end{small}
    \vspace{-10pt}
  \end{wraptable}

\para{Generation quality (MAUVE).}
\Cref{tab:mauve} reports MAUVE scores~\citep{pillutla2021mauve} for Dream, measuring distributional similarity between student-generated and reference text. All on-policy variants substantially improve over the off-policy baseline of $0.60$. Interestingly, the unconstrained trajectory variant with $K=4$ achieves the highest MAUVE score of $0.76$, while adding confidence weighting and the trust-region anchor slightly reduces MAUVE. This suggests a precision--diversity trade-off: CW and TR improve token-level accuracy by down-weighting uncertain teacher targets and stabilizing updates, but they also modestly reduce distributional coverage.

\subsection{Ablations}
\label{sec:ablation}

\para{Choice of $\alpha$ and $\beta$ for Off-Policy $\alpha$-$\beta$ divergence on MDLM.}
We sweep $(\alpha,\beta)$ for training MDLM and compare the perplexity once all configurations have converged. Although \citet{wang2025abkd} report that small $\alpha$ and large $\beta$ are best for AR distillation, we observe a different regime here: small $\alpha\!=\!0.2$ paired with moderate $\beta\!=\!0.5$ is best (\Cref{tab:offpolicy_abkl}). We adopt this configuration as the off-policy initialization and reuse it during the on-policy phase.

\para{Number of Trajectory Timesteps $K$ for On-Policy MDLM.}
\Cref{tab:onpolicy_trajactory} sweeps $K\in\{1,4,6,8\}$ and compares against reverse-KL on-policy distillation. $K\!=\!4$ is best; larger values over-constrain the student and degrade perplexity. Notably, trajectory-consistent $\alpha$-$\beta$ divergence at $K\!=\!4$  surpasses the teacher's own perplexity, supporting the view that intermediate-state supervision provides a richer signal than final-output matching.

\begin{figure}[h]
  \centering
  \includegraphics[width=0.8\columnwidth]{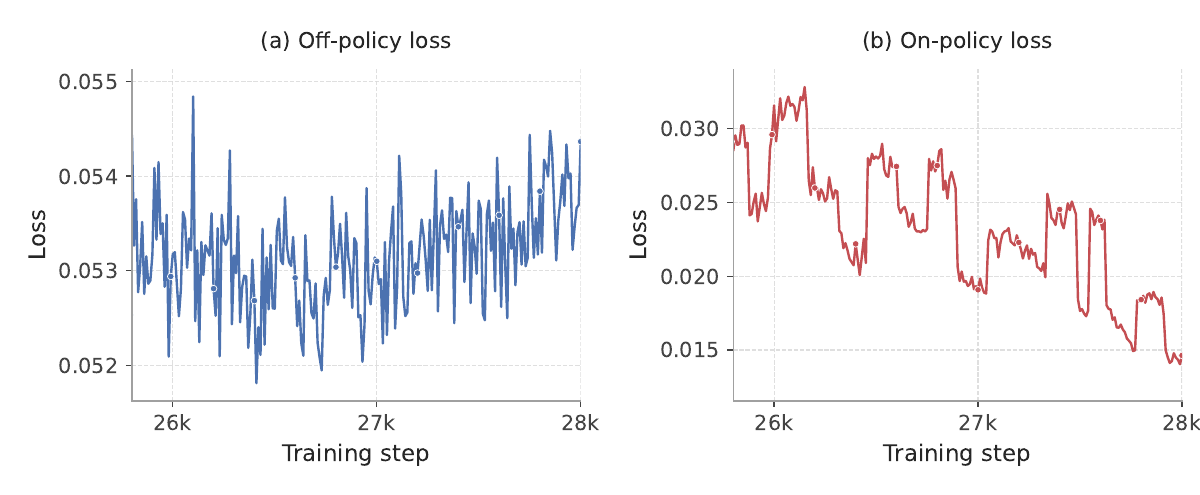}
  \caption{\textbf{Loss dynamics under linear curriculum mixing for Dream.} (a) The off-policy loss decreases initially and then rises. (b) The on-policy loss shows abrupt rises and drops despite an overall downward trend.}
  \label{fig:curriculum_linear_losses}
\end{figure}

\begin{table}[t]
\centering
\begin{minipage}[t]{0.38\textwidth}
  \centering
  \caption{Off-policy MDLM perplexity for different $(\alpha,\beta)$ in $\alpha$-$\beta$ divergence.}
  \label{tab:offpolicy_abkl}
  \begin{small}
  \setlength{\tabcolsep}{12pt}
  \begin{tabular}{@{}ccc@{}}
  \toprule
  $\alpha$ & $\beta$ & GPT-2 PPL $\downarrow$ \\
  \midrule
  0.2 & 0.2 & 65.75 \\
  0.2 & 0.5 & \textbf{32.48} \\
  0.2 & 0.7 & 59.98 \\
  0.2 & 0.9 & 150.91 \\
  0.5 & 0.5 & 88.75 \\
  0.7 & 0.2 & 89.99 \\
  0.7 & 0.7 & 530.22 \\
  \bottomrule
  \end{tabular}
  \end{small}
\end{minipage}
\hfill
\begin{minipage}[t]{0.59\textwidth}
  \centering
  \caption{Effect of the number of trajectory timesteps $K$ on MDLM on-policy distillation. Lower is better.}
  \label{tab:onpolicy_trajactory}
  \begin{small}
  \setlength{\tabcolsep}{20pt}
  \begin{tabular}{@{}lc@{}}
  \toprule
  Method & GPT-2 PPL $\downarrow$ \\
  \midrule
  $K=1$ & 27.64 \\
  $K=4$ & \textbf{25.38} \\
  $K=6$ & 25.80 \\
  $K=8$ & 26.78 \\
  \midrule
  Reverse-KL on-policy & 27.72 \\
  Teacher reference & 27.04 \\
  \bottomrule
  \end{tabular}
  \end{small}
\end{minipage}
\end{table}
\para{Curriculum on the on-policy weight does not transfer from AR.} \citet{agarwal2024policy} use a mixture of real data and student-generated samples for on-policy AR LLM distillation. We tried the same idea in our setting by gradually mixing on-policy data into off-policy training with a stepwise / linear / cosine schedule on the on-policy weight $h$. This results in model collapse for all three schedules. For Dream, \Cref{fig:curriculum_linear_losses} shows the loss curves under the linear schedule. The off-policy loss initially decreases and then increases, while the on-policy loss exhibits abrupt rises and drops despite an overall downward trend. These trends suggest competing objectives and unstable optimization. We hypothesize that the conflict arises from their different state distributions: off-policy distillation matches the teacher on forward-noised teacher samples, whereas on-policy distillation matches it on states visited by the student's reverse process. Joint optimization may therefore destabilize the student and contribute to collapse. This forward--reverse mismatch does not exist in AR models. We therefore use a hard switch from off-policy to on-policy in the final framework. This shows that AR and DLM training dynamics differ in non-trivial ways, and that AR-derived recipes do not always carry over.

\subsection{Zero-shot benchmarks}
\label{sec:zeroshot}
\Cref{tab:zeroshot} reports zero-shot accuracy on HellaSwag (HS), PIQA, ARC-Easy (ARC-E), and Winogrande (WG) for the Dream student, starting from the optimal $300$M-token off-policy checkpoint. On-policy trajectory training preserves the benchmark performance of the off-policy model and slightly improves several metrics, including PIQA and ARC-Easy. Adding confidence reweighting keeps HS, PIQA, and WG broadly comparable, although ARC-Easy decreases modestly. Overall, the zero-shot results remain close across ablations, while \Cref{tab:main} shows substantial perplexity improvements. Thus, within the capacity of a smaller student, our framework transfers the teacher’s generation behavior more effectively than forward-KL, improves generation quality, and preserves downstream performance.

\begin{table}[h]
\centering
  \caption{Zero-shot accuracy of the Dream-93M student on-policy ablations starting from the 300M off-policy checkpoint.}
  \label{tab:zeroshot}
  \setlength{\tabcolsep}{8pt}
  \begin{tabular}{@{}lcccc@{}}
  \toprule
  Setting & HS & PIQA & ARC-E & WG \\
  \midrule
  off-policy ckpt & 0.2742 & 0.5582 & 0.3481 & 0.4949 \\
  traj.\ $K\!=\!4$ & \textbf{0.2751} & \textbf{0.5609} & \textbf{0.3514} & 0.4941 \\
  traj.\ $K\!=\!4$ + CW & 0.2723 & 0.5566 & 0.3308 & \textbf{0.4957} \\
  traj.\ $K\!=\!4$ + CW + TR & 0.2719 & 0.5566 & 0.3316 & 0.4949 \\
  \bottomrule
  \end{tabular}
\end{table}

\subsection{Qualitative samples}
We provide qualitative MDLM samples in \Cref{app:samples}, comparing the FKL-distilled student and our student. The FKL student exhibits severe degeneration, including repetition loops, malformed phrases, and weak discourse structure. Our student substantially mitigates these issues, producing longer stretches with a more stable topic and fewer catastrophic loops under the same sampling budget.

\section{Conclusion}
\label{sec:conclusion}

We presented the first systematic study of knowledge distillation for diffusion language models, identified two failure modes of classical forward-KL distillation, and proposed a unified data-free distillation framework that combines off- and on-policy phases, a trajectory loss along the student's own reverse process, confidence reweighting, and a trust-region anchor. The resulting students substantially improve over FKL baselines in both the small (MDLM-110M) and large (Dream-7B) regimes, with the MDLM student surpassing its teacher in perplexity and the Dream student preserving zero-shot accuracy at the smaller student scale. We hope this serves as a foundation for further work on scalable training and sampling for diffusion-based language models.

\section*{Acknowledgments}
M. Rodrigues is supported by the Engineering and Physical Sciences Research Council (EPSRC) through the AI Hub in Generative Models [grant
number EP/Y028805/1].
S. Yoon was supported in part by the InnoCORE program of the Ministry of Science and ICT (No. 1.260017.01), the Institute of Information \& Communications Technology Planning \& Evaluation (IITP) grants funded by the Korea government (MSIT) (RS-2025-25442824, AI Star Fellowship Program (Ulsan National Institute of Science and Technology)) and (No. RS-2020-II201336, Artificial Intelligence Graduate School Program (UNIST)), and the National Research Foundation of Korea (NRF) grant funded by the Korea government (MSIT) (No. RS-2026-25506577).

\bibliographystyle{plainnat}
\bibliography{example_paper}

%%%%%%%%%%%%%%%%%%%%%%%%%%%%%%%%%%%%%%%%%%%%%%%%%%%%%%%%%%%%

\newpage
\appendix
\onecolumn

\section{Extended Theoretical Results}
\label{app:theory}

\subsection{Telescoping error bound}
\label{app:telescoping}

\begin{theorem}[On-Policy Error Decomposition]
\label{thm:tv-decomposition}
Let $p$ and $q_\theta$ denote the joint distributions over reverse trajectories induced by the teacher and student, respectively. Both chains start from the fully masked state $x_T \sim \mu$. Then
\begin{equation}
\mathrm{TV}\!\bigl(q_\theta(x_0),\; p(x_0)\bigr) \;\leq\; \sum_{t=1}^{T} \,\mathbb{E}_{x_t \sim q_\theta(\cdot, t)}\!\Big[\mathrm{TV}\!\bigl(p_\theta(\cdot \mid x_t, t),\; p_{T}(\cdot \mid x_t, t)\bigr)\Big],
\end{equation}
where $q_\theta(\cdot, t)$ is the marginal under the student's reverse process.
\end{theorem}

\begin{proof}
Define the hybrid chain $q^{(s)}$ that uses the student kernel for steps $T, \ldots, s+1$ and the teacher kernel for steps $s, \ldots, 1$. Then $q^{(0)} = q_\theta$ and $q^{(T)} = p$. By the triangle inequality on marginals at $x_0$:
\begin{equation}
\mathrm{TV}\!\bigl(q_\theta(x_0),\; p(x_0)\bigr) \;\leq\; \sum_{s=1}^{T} \mathrm{TV}\!\bigl(q^{(s-1)}(x_0),\; q^{(s)}(x_0)\bigr).
\end{equation}
The chains $q^{(s-1)}$ and $q^{(s)}$ differ only at step $s$ and share the upstream marginal $q_\theta(\cdot, s)$. By the data processing inequality (post-processing by the shared teacher kernels for steps $s-1, \ldots, 1$ cannot increase TV):
\begin{equation}
\mathrm{TV}\!\bigl(q^{(s-1)}(x_0),\; q^{(s)}(x_0)\bigr) \;\leq\; \mathbb{E}_{x_s \sim q_\theta(\cdot, s)}\!\Big[\mathrm{TV}\!\bigl(p_\theta(\cdot \mid x_s, s),\; p_{T}(\cdot \mid x_s, s)\bigr)\Big].
\end{equation}
Summing over $s$ yields the result.
\end{proof}

\begin{corollary}[Trajectory Loss as Unbiased Estimator]
\label[corollary]{cor:traj-loss}
Let $x_{0:T} \sim q_\theta$ be a student reverse trajectory, and let $t_1,\dots,t_K$ be drawn i.i.d.\ uniformly from $\{1,\dots,T\}$, independently of the trajectory (sorting the $t_k$ does not affect the result). Define $D_t(x_t) \coloneqq D(p_T(\cdot \mid x_t, t)\,\Vert\,p_\theta(\cdot \mid x_t, t))$ and assume $\mathbb{E}_{x_t \sim q_\theta}[D_t] < \infty$ for every $t$. Then
\begin{equation}
\mathbb{E}\!\left[\mathcal{L}_{\mathrm{traj}}(\theta)\right] \;=\; \frac{1}{T}\sum_{t=1}^{T}\mathbb{E}_{x_t \sim q_\theta(\cdot, t)}\!\left[D_t(x_t)\right] \;=:\; \overline{D}.
\end{equation}
Consequently, when $D = \mathrm{KL}$ and $p_\theta$ has full support, combining \Cref{thm:tv-decomposition} with Pinsker's inequality and the Cauchy--Schwarz inequality yields
\begin{equation}
\mathrm{TV}\!\bigl(q_\theta(x_0),\,p(x_0)\bigr) \;\leq\; \sum_{t=1}^{T} \mathbb{E}_{x_t \sim q_\theta}\!\left[\sqrt{\tfrac{1}{2}D_t}\right] \;\leq\; \sqrt{T}\,\sqrt{\tfrac{1}{2}\textstyle\sum_t \mathbb{E}[D_t]} \;=\; T\sqrt{\tfrac{1}{2}\,\overline{D}}.
\end{equation}
Hence minimizing $\mathcal{L}_{\mathrm{traj}}$ in expectation minimizes an upper bound on $\mathrm{TV}(q_\theta(x_0), p(x_0))$, with explicit constant $T\sqrt{\overline{D}/2}$.
\end{corollary}

\begin{proof}
For each $k$, by independence of $t_k$ from the trajectory and uniformity of $t_k$ on $\{1,\dots,T\}$,
\begin{equation}
\mathbb{E}\!\left[D_{t_k}(x_{t_k})\right] \;=\; \frac{1}{T}\sum_{t=1}^{T}\mathbb{E}_{x_t \sim q_\theta(\cdot,t)}\!\left[D_t(x_t)\right] \;=\; \overline{D}.
\end{equation}
Linearity of expectation over $k=1,\dots,K$ then gives $\mathbb{E}[\mathcal{L}_{\mathrm{traj}}] = \overline{D}$, proving the first claim. For the TV bound, \Cref{thm:tv-decomposition} gives $\mathrm{TV}(q_\theta(x_0),p(x_0)) \leq \sum_t \mathbb{E}_{x_t \sim q_\theta}[\mathrm{TV}_t]$. Pinsker's inequality applied pointwise yields $\mathrm{TV}_t \leq \sqrt{D_t/2}$, and Jensen's inequality on the concave $\sqrt{\cdot}$ gives $\mathbb{E}[\sqrt{D_t/2}] \leq \sqrt{\mathbb{E}[D_t]/2}$. Finally, Cauchy--Schwarz on $\sum_t a_t \leq \sqrt{T}\sqrt{\sum_t a_t^2}$ with $a_t = \sqrt{\mathbb{E}[D_t]/2}$ produces the factor $T$.
\end{proof}

\subsection{Proof of \Cref{prop:offpolicy_fail} (Off-policy failure)}
\label{app:offpolicy_fail}

\begin{proof}
Fix vocabulary $\mathcal{V}=\{a,b\}$, sequence length $n$, and $T=2$. Let $p_{\mathrm{data}} = \frac{1}{2}\delta_{(a,\ldots,a)} + \frac{1}{2}\delta_{(b,\ldots,b)}$. The forward process masks each token independently with probability $\frac{1}{2}$ at $t=1$ and fully masks at $t=2$.

\emph{Zero off-policy loss.}
Recall the data distribution $p_{\mathrm{data}} = \tfrac{1}{2}\delta_{(a,\ldots,a)} + \tfrac{1}{2}\delta_{(b,\ldots,b)}$,
so any clean sample $x_0$ is \emph{monochromatic}: either all-$a$ or all-$b$.
Off-policy training visits states $x_t$ obtained by forward-noising such an $x_0$, i.e.
$x_t \sim q(\cdot \mid x_0)$ with $x_0 \sim p_{\mathrm{data}}$.

\textbf{(1) What states are visited at $t=1$.}
At $t=1$ each position is independently masked with probability $\tfrac{1}{2}$, so
a typical visited state $x_1$ has some subset $U\subseteq\{1,\dots,n\}$ of unmasked positions
and the remaining positions equal to $\mathtt{[MASK]}$. Crucially, because $x_0$ is monochromatic,
\emph{every} unmasked position in $x_1$ shows the \emph{same} symbol $s\in\{a,b\}$.
Mixed states such as $(a,b,\mathtt{M},a,\dots)$ are \textbf{never} visited by the forward
process---they have probability zero under $q(x_1)$. This is the off-policy
support that the student is trained on.

\textbf{(2) The true per-position posterior on this support.}
Fix any visited $x_1$ with at least one unmasked position $j\in U$, all showing symbol $s$.
For any masked position $i\notin U$, Bayes' rule gives
\begin{equation*}
p(x_0^i = c \mid x_1) \;=\; \frac{\sum_{x_0}\mathbf{1}[x_0^i=c]\,p_{\mathrm{data}}(x_0)\,q(x_1\mid x_0)}
{\sum_{x_0}p_{\mathrm{data}}(x_0)\,q(x_1\mid x_0)}.
\end{equation*}
Only $x_0=(s,\ldots,s)$ is consistent with $x_1$ (the other monochromatic candidate
$(\bar s,\ldots,\bar s)$ is ruled out by the unmasked $j$), so the posterior collapses
to the point mass $p(x_0^i = c\mid x_1) = \delta_{c=s}$.
At $t=2$ the entire sequence is masked, no unmasked position constrains $x_0$, and the
posterior is the uniform mixture $\tfrac{1}{2}\delta_a + \tfrac{1}{2}\delta_b$ at every position.

\textbf{(3) The student that achieves zero off-policy loss.}
Define a per-position student
\begin{equation*}
p_\phi(x_0^i = c \mid x_t, t) \;=\;
\begin{cases}
\delta_{c=s(x_t)}, & t=1 \text{ and } x_t \text{ has any unmasked position showing } s(x_t),\\
\tfrac{1}{2}\delta_a + \tfrac{1}{2}\delta_b, & t=2,\\
\text{arbitrary}, & \text{on states outside the forward support.}
\end{cases}
\end{equation*}
On every state in $\mathrm{supp}\,q$, $p_\phi$ matches the true posterior \emph{exactly},
so the off-policy distillation loss
$\mathbb{E}_{t,\,x_t\sim q}\bigl[D\!\left(p_T(\cdot\mid x_t,t)\,\Vert\,p_\phi(\cdot\mid x_t,t)\right)\bigr]$
is identically zero, regardless of the divergence $D$.
The behavior of $p_\phi$ on \emph{off-support} states (e.g.\ mixed-symbol $x_1$)
is unconstrained by the off-policy objective---and this is precisely the freedom
that the next paragraph exploits to make generation fail.

\emph{Generation fails.}
We now show that the very same student $p_\phi$ that achieves zero off-policy loss
generates valid monochromatic sequences with vanishing probability.

\textbf{(1) The student's reverse process.}
Sampling starts from $x_2 = (\mathtt{M},\ldots,\mathtt{M})$, the fully masked state.
At $t=2$ the student's per-position posterior is $\tfrac{1}{2}\delta_a + \tfrac{1}{2}\delta_b$.
We adopt the standard masked-DLM denoising step: at each reverse step a fresh subset of masked
positions is unmasked by sampling \emph{independently} from the student's per-position
posterior at those positions. Per-position independence at a single denoising step is the assumption made by virtually all masked-DLM samplers \citep{sahoo2024simple,nie2025large,ye2025dream} and is exactly the assumption \citet{hayakawa2025distillation} identify as the source of difficulty in few-step distillation.

In our $T=2$ construction, after the first reverse step the chain transitions
$x_2 \to x_1$, where each position is, with probability $\tfrac{1}{2}$, unmasked
to an independent uniform draw from $\{a,b\}$ (and otherwise stays masked). The
second step $x_1 \to x_0$ then fills the remaining masked positions according to
$p_\phi(\cdot\mid x_1, 1)$.

\textbf{(2) Probability of landing in the forward support after step 1.}
Let $U \subseteq \{1,\dots,n\}$ be the set of positions unmasked at step 1, and let
$k = |U|$. Conditioned on $U$, each unmasked position is an i.i.d.\ uniform draw
from $\{a,b\}$, so the probability that all $k$ unmasked positions show the same
symbol is
\begin{equation*}
\Pr[\text{$x_1$ monochromatic on $U$} \mid |U|=k] \;=\;
\begin{cases} 1, & k\leq 1,\\ 2^{1-k}, & k\geq 1. \end{cases}
\end{equation*}
Marginalizing over $|U|\sim\mathrm{Binomial}(n,\tfrac{1}{2})$,
\begin{align*}
\Pr[\text{$x_1$ monochromatic on $U$}]
&\;=\; \sum_{k=0}^{n} \binom{n}{k}\!\left(\tfrac{1}{2}\right)^{\!n}\!\cdot 2^{1-\max(k,1)} \\
&\;\leq\; \sum_{k=0}^{n} \binom{n}{k}\!\left(\tfrac{1}{2}\right)^{\!n}\!\cdot 2\cdot 2^{-k}
\;=\; 2\left(\tfrac{3}{4}\right)^{\!n},
\end{align*}
where the last equality uses the binomial identity
$\sum_k \binom{n}{k}(1/2)^n 2^{-k} = (1/2 + 1/4)^n = (3/4)^n$.
Equivalently: for the student to even \emph{enter} the forward support at step 1,
all $|U|$ independent fair-coin draws must agree, an event of probability
exponentially small in $n$.

\textbf{(3) Why this implies generation fails.}
Conditioned on $x_1$ being monochromatic on $U$ (showing some $s\in\{a,b\}$),
the student's step-1 posterior $p_\phi(\cdot\mid x_1,1)=\delta_s$ correctly fills
the remaining masked positions with $s$, producing a valid monochromatic $x_0$.
Conditioned on $x_1$ being mixed-symbol (off the forward support), the student's
behavior is unconstrained by the off-policy objective. Hence
\begin{equation*}
\Pr_{x_0\sim p_\phi^{\mathrm{gen}}}\!\bigl[x_0 \in \{(a,\ldots,a),(b,\ldots,b)\}\bigr]
\;\leq\; \Pr[\text{$x_1$ monochromatic on $U$}]
\;\leq\; 2\left(\tfrac{3}{4}\right)^{\!n}.
\end{equation*}

\textbf{(4) From mass on the support to TV.}
Since $p_{\mathrm{data}}$ places all of its mass on the two monochromatic
sequences and $p_\phi^{\mathrm{gen}}$ places at most $2(3/4)^n$ there,
\begin{equation*}
\mathrm{TV}(p_\phi^{\mathrm{gen}}, p_{\mathrm{data}})
\;\geq\; p_{\mathrm{data}}\bigl(\{(a,\ldots,a),(b,\ldots,b)\}\bigr) - p_\phi^{\mathrm{gen}}\bigl(\{(a,\ldots,a),(b,\ldots,b)\}\bigr)
\;\geq\; 1 - 2\left(\tfrac{3}{4}\right)^{\!n}.
\end{equation*}
For any $\epsilon>0$, choosing
$n \geq \log\!\bigl(2/\epsilon\bigr)/\log(4/3)$
makes $2(3/4)^n \leq \epsilon$, so $\mathrm{TV}(p_\phi^{\mathrm{gen}}, p_{\mathrm{data}}) > 1-\epsilon$,
as claimed.

\end{proof}

\begin{lemma}[Pinsker-type control for $\alpha$-$\beta$ divergence]
\label[lemma]{lem:abkl-pinsker}
Let $P$ and $Q$ be probability distributions on a finite alphabet. For $\alpha>0$, $\beta>0$, and $s\coloneqq\alpha+\beta\leq 1$, the $\alpha$-$\beta$ divergence satisfies
\begin{equation}
\mathrm{TV}(P,Q)
\;\leq\;
\sqrt{\frac{1}{2}\operatorname{D}_{\alpha,\beta}(P\Vert Q)}.
\label{eq:abkl-pinsker}
\end{equation}
\end{lemma}

\begin{proof}
Let $\lambda=\alpha/s\in(0,1)$. Write
\begin{equation}
G_s(P,Q)
\;=\;
\sum_c
\left[
\lambda P(c)^s + (1-\lambda)Q(c)^s
- P(c)^{s\lambda}Q(c)^{s(1-\lambda)}
\right],
\end{equation}
so that
\begin{equation}
\operatorname{D}_{\alpha,\beta}(P\Vert Q)
\;=\;
\frac{G_s(P,Q)}{s^2\lambda(1-\lambda)}.
\end{equation}
For each coordinate with positive mass, set $u=\log P(c)$ and $v=\log Q(c)$; zero probabilities are handled by continuity. The coordinate summand in $G_s$ is the Jensen gap of $\varphi_s(z)=e^{sz}$ between $u$ and $v$ with weights $\lambda$ and $1-\lambda$. Since $u,v\leq 0$ and $s\leq 1$,
\begin{equation}
\left(\varphi_s - s^2\varphi_1\right)''(z)
\;=\;
s^2\left(e^{sz}-e^z\right)
\;\geq\; 0
\qquad \text{for all } z\leq 0.
\end{equation}
Thus the Jensen gap of $\varphi_s$ is at least $s^2$ times the Jensen gap of $\varphi_1$, and therefore
\begin{equation}
G_s(P,Q) \;\geq\; s^2 G_1(P,Q).
\end{equation}
It follows that
\begin{equation}
\operatorname{D}_{\alpha,\beta}(P\Vert Q)
\;\geq\;
\frac{G_1(P,Q)}{\lambda(1-\lambda)}
\;=\;
\frac{1-\sum_c P(c)^\lambda Q(c)^{1-\lambda}}{\lambda(1-\lambda)}.
\end{equation}
The right-hand side is the normalized alpha divergence generated by
\begin{equation}
f_\lambda(r)
\;=\;
\frac{\lambda r + (1-\lambda) - r^\lambda}{\lambda(1-\lambda)},
\qquad
f_\lambda''(1)=1.
\end{equation}
By the standard Pinsker inequality for this $f$-divergence \citep[Theorem~3]{gilardoni2008pinsker},
\begin{equation}
\frac{1-\sum_c P(c)^\lambda Q(c)^{1-\lambda}}{\lambda(1-\lambda)}
\;\geq\;
2\,\mathrm{TV}(P,Q)^2.
\end{equation}
Combining the two inequalities gives \Cref{eq:abkl-pinsker}.
\end{proof}

\subsection{Proof of \Cref{thm:main} (Unified distillation bound)}
\label{app:main_thm}

\begin{proof}
The argument has three steps.

\emph{Step 1: Telescoping.} Since $\ell \in [0, B]$, the variational representation of TV gives $\bigl|\mathbb{E}_{q_\theta}[\ell] - \mathbb{E}_p[\ell]\bigr| \leq B \cdot \mathrm{TV}\bigl(q_\theta(x_0), p(x_0)\bigr)$. Applying \Cref{thm:tv-decomposition}:
\begin{equation*}
\bigl|\mathbb{E}_{q_\theta}[\ell] - \mathbb{E}_p[\ell]\bigr| \;\leq\; B\cdot\mathrm{TV}\bigl(q_\theta(x_0), p(x_0)\bigr) \;\leq\; B \sum_{t=1}^{T} \mathbb{E}_{x_t \sim q_\theta}[\mathrm{TV}(p_T \Vert p_\theta \mid x_t)].
\end{equation*}

\emph{Step 2: TV calibration  $+$ Jensen $+$ Cauchy--Schwarz.} We convert the sum of TV's into a single $\sqrt{\overline{D}}$ expression in three sub-steps. Write $D_t(x_t) \coloneqq D(p_T(\cdot|x_t,t) \Vert p_\theta(\cdot|x_t,t))$ and $a_t \coloneqq \mathbb{E}_{x_t \sim q_\theta}[D_t]$.

\emph{(a)} By the calibration condition in \Cref{eq:calibrated-divergence}, for each fixed $x_t$, $\mathrm{TV}(p_T \Vert p_\theta \mid x_t) \leq \kappa_D\sqrt{D_t(x_t)}$.

\emph{(b) Jensen on the inner expectation.} Since $\sqrt{\cdot}$ is concave, $\mathbb{E}_{x_t}[\kappa_D\sqrt{D_t}] \leq \kappa_D\sqrt{\mathbb{E}_{x_t}[D_t]} = \kappa_D\sqrt{a_t}$.

\emph{(c) Cauchy--Schwarz on the sum over $t$.} With $u_t = 1$ and $v_t = \sqrt{a_t}$:
\begin{equation*}
\sum_{t=1}^T \sqrt{a_t} \;=\; \sum_t u_t v_t \;\leq\; \sqrt{\textstyle\sum_t u_t^2}\,\sqrt{\textstyle\sum_t v_t^2} \;=\; \sqrt{T}\,\sqrt{\textstyle\sum_t a_t} \;=\; T\sqrt{\overline{D}},
\end{equation*}
where $\overline{D} \coloneqq \tfrac{1}{T}\sum_t a_t$ is the time-averaged divergence. Combining (a)--(c):
\begin{equation*}
B \sum_t \mathbb{E}_{x_t \sim q_\theta}[\mathrm{TV}_t]
\;\leq\; B \kappa_D\sum_t \sqrt{a_t}
\;\leq\; BT \kappa_D\,\sqrt{\overline{D}}.
\end{equation*}

\emph{Step 3: CW splitting.} The trajectory loss is evaluated under the current rollout $\rho_\theta$ (Algorithm~\ref{alg:onpolicy}, Line~1), so $\overline{D} = \tfrac{1}{T}\sum_t \mathbb{E}_{x_t \sim q_\theta}[D_t]$ is exactly the (unweighted) trajectory loss in expectation. Decompose pointwise via the confidence weight: $D_t = w_t\,D_t + (1-w_t)\,D_t$. Taking expectations and averaging over $t$:
\begin{equation*}
\overline{D} \;=\; \mathcal{L}_{\mathrm{traj}}^{\mathrm{CW}} \;+\; \varepsilon_{\mathrm{lowconf}},
\end{equation*}
where $\mathcal{L}_{\mathrm{traj}}^{\mathrm{CW}} \coloneqq \tfrac{1}{T}\sum_t \mathbb{E}_{x_t \sim q_\theta}[w_t\,D_t]$ and $\varepsilon_{\mathrm{lowconf}} \coloneqq \tfrac{1}{T}\sum_t \mathbb{E}_{x_t \sim q_\theta}[(1-w_t)\,D_t]$. Invoking the teacher-calibration assumption \eqref{eq:lowconf-bound}, $\varepsilon_{\mathrm{lowconf}}\leq\varepsilon_0$, so $\overline{D}\leq\mathcal{L}_{\mathrm{traj}}^{\mathrm{CW}}+\varepsilon_0$. Combining with Step~2 yields
\begin{equation}
\bigl|\mathbb{E}_{q_\theta}[\ell] - \mathbb{E}_p[\ell]\bigr| \;\leq\; BT\kappa_D\sqrt{\mathcal{L}_{\mathrm{traj}}^{\mathrm{CW}} + \varepsilon_0}.
\end{equation}

\emph{Remark on the trust region.} The trust-region penalty $\mathcal{L}_{\mathrm{TR}}$ does \emph{not} appear in this bound: it is an optimization stabilizer that keeps $\theta$ close to a recently known good policy $\theta_{\mathrm{old}}$ between refreshes, controlling the non-stationarity of $\rho_\theta$ during training rather than tightening the generation-quality bound. Its role is analogous to the trust region in TRPO~\citep{schulman2015trust}: it bounds how much the rollout distribution can shift between consecutive optimizer steps, ensuring the per-step gradient estimate remains a faithful proxy for the true on-policy gradient (\Cref{app:tr_convergence}).
\end{proof}

\subsection{CW gradient variance reduction}
\label{app:cw_variance}

\begin{proposition}[CW Variance Reduction]
\label[proposition]{prop:cw-variance}
Let $g_i$ be per-position gradient vectors with confidence weight $w_i = (\max_c p_T(c\mid x_t,t,i))^\gamma$. Partition positions into $\mathcal{S}_{\mathrm{on}} = \{i: \max_c p_T \geq \tau\}$ and $\mathcal{S}_{\mathrm{off}} = \mathcal{I}\setminus\mathcal{S}_{\mathrm{on}}$. Under conditional uncorrelatedness of per-position gradients, let $\alpha$ be the fraction of total gradient variance from off-manifold positions. Then:
\begin{equation}
\frac{\mathrm{tr}\,\mathrm{Var}[G^{\mathrm{CW}}]}{\mathrm{tr}\,\mathrm{Var}[G]} \;\leq\; 1 - \alpha(1-\tau^{2\gamma}).
\end{equation}
When off-manifold variance dominates ($\alpha\to 1$), the ratio approaches $\tau^{2\gamma}$, vanishing exponentially in $\gamma$.
\end{proposition}

\begin{proof}
Under conditional uncorrelatedness: $\mathrm{Var}[G^{\mathrm{CW}}] = n^{-2}\sum_i w_i^2\,\mathrm{Var}[g_i]$. For $i\in\mathcal{S}_{\mathrm{on}}$: $w_i^2 \leq 1$. For $i\in\mathcal{S}_{\mathrm{off}}$: $w_i^2 < \tau^{2\gamma}$. Dividing by $\mathrm{Var}[G] = n^{-2}\sum_i \mathrm{Var}[g_i]$ gives $\frac{(1-\alpha)+\tau^{2\gamma}\alpha}{1} = 1 - \alpha(1-\tau^{2\gamma})$.
\end{proof}

\subsection{Trust-region anchor bounds distribution shift}
\label{app:tr_convergence}

\begin{proposition}
\label[proposition]{prop:trust-region}
If the trust-region constraint $\mathbb{E}_{t, x_t\sim\rho_{\theta'}}[\mathrm{KL}(p_{\theta_{\mathrm{old}}}\Vert p_{\theta'})] \leq \varepsilon$ holds and the per-step loss is bounded by $B_\ell$, then:
\begin{equation}
\bigl|\mathcal{L}_{\mathrm{traj}}(\theta') - \mathcal{L}_{\mathrm{traj}}^{\rho_{\theta_{\mathrm{old}}}}(\theta')\bigr|
\;\leq\; B_\ell \cdot T\sqrt{\varepsilon/2},
\end{equation}
where $\mathcal{L}_{\mathrm{traj}}^{\rho_{\theta_{\mathrm{old}}}}$ is the loss under the old rollout distribution. 
\end{proposition}

\begin{proof}
The argument adapts the TRPO performance-difference bound~\citep[Theorem 1]{schulman2015trust} to the finite-horizon masked-DLM setting without discount factor.

\emph{Step 1 (Expectation gap by TV).} Write $\ell_t(x_t) \in [0, B_\ell]$ for the per-step loss at time $t$. The variational representation of TV gives, for each $t$,
\begin{equation*}
\bigl|\mathbb{E}_{x_t \sim \rho_{\theta'}}[\ell_t] - \mathbb{E}_{x_t \sim \rho_{\theta_{\mathrm{old}}}}[\ell_t]\bigr| \;\leq\; B_\ell \cdot \mathrm{TV}\bigl(\rho_{\theta'}(\cdot,t),\, \rho_{\theta_{\mathrm{old}}}(\cdot,t)\bigr).
\end{equation*}

\emph{Step 2 (State-marginal TV via second telescoping).} The two rollouts share the initial state $x_T$ and differ only in their per-step kernels $p_{\theta'}$ vs.\ $p_{\theta_{\mathrm{old}}}$. By the same telescoping argument as \Cref{thm:tv-decomposition},
\begin{equation*}
\mathrm{TV}\bigl(\rho_{\theta'}(\cdot,t),\, \rho_{\theta_{\mathrm{old}}}(\cdot,t)\bigr) \;\leq\; \sum_{s=t+1}^{T} \mathbb{E}_{x_s \sim \rho_{\theta'}(\cdot,s)}\!\bigl[\mathrm{TV}(p_{\theta_{\mathrm{old}}} \Vert p_{\theta'} \mid x_s)\bigr].
\end{equation*}

\emph{Step 3 (Pinsker on per-step TV).} For each $s$, $\mathrm{TV}(p_{\theta_{\mathrm{old}}} \Vert p_{\theta'} \mid x_s) \leq \sqrt{D(p_{\theta_{\mathrm{old}}} \Vert p_{\theta'} \mid x_s)/2}$. Combined with Jensen's inequality on the inner expectation and Cauchy--Schwarz on the sum over $s$:

\begin{equation*}
\sum_{s=t+1}^{T} \mathbb{E}_{\rho_{\theta'}}\!\bigl[\sqrt{\mathrm{KL}_s^{\mathrm{TR}}/2}\bigr]
\;\leq\; \sum_{s=t+1}^{T} \sqrt{\mathbb{E}_{\rho_{\theta'}}[\mathrm{KL}_s^{\mathrm{TR}}]/2}
\;\leq\; \sqrt{T}\,\sqrt{\textstyle\sum_{s=1}^{T} \mathbb{E}_{\rho_{\theta'}}[\mathrm{KL}_s^{\mathrm{TR}}]/2}
\;\leq\; T \sqrt{\varepsilon/2},
\end{equation*}
using $\mathbb{E}_{\rho_{\theta'}}[\mathrm{KL}_s^{\mathrm{TR}}] \leq \varepsilon$ uniformly in $s$ from the constraint. So $\mathrm{TV}(\rho_{\theta'}(\cdot,t), \rho_{\theta_{\mathrm{old}}}(\cdot,t)) \leq T\sqrt{\varepsilon/2}$ for each $t$.

\emph{Step 4 (Average over $t$).} Combining Steps~1--3 and averaging over $t$:
\begin{equation*}
\bigl|\mathcal{L}_{\mathrm{traj}}(\theta') - \mathcal{L}_{\mathrm{traj}}^{\rho_{\theta_{\mathrm{old}}}}(\theta')\bigr|
\;\leq\; B_\ell \cdot T\sqrt{\varepsilon/2}.
\end{equation*}
\end{proof}

\subsection{Optimal $K$: bias--variance tradeoff}
\label{app:kstep}

\begin{proposition}[Gradient Variance Decomposition]
\label[proposition]{prop:bias-variance}
Let $\sigma^2$ be the average per-timestep gradient variance and $\bar\rho$ the average pairwise gradient correlation across timesteps from the same trajectory. Then:
\begin{equation}
\mathrm{tr}\bigl(\mathrm{Cov}[\nabla\mathcal{L}_K]\bigr) \;\approx\; \frac{(1-\bar\rho)\,\sigma^2}{K} + \bar\rho\,\sigma^2.
\end{equation}
The first term is reducible by increasing $K$; the second is an irreducible floor from intra-trajectory correlation.
\end{proposition}

\begin{corollary}[Compute-optimal $K$ under a fixed budget]
\label[corollary]{cor:compute-optimal-k}
Assume a total compute budget $B_{\mathrm{comp}}$.
Each independent trajectory costs $C_{\mathrm{traj}}$, and querying/scoring each sampled timestep costs $c$.
Thus the number of independent trajectories that can be used with $K$ sampled timesteps is approximately
\begin{equation}
N(K)
=
\frac{B_{\mathrm{comp}}}{C_{\mathrm{traj}}+cK}.
\end{equation}
Under the variance model of \Cref{prop:bias-variance}, the variance of the averaged gradient estimator scales as
\begin{equation}
\mathrm{Var}_{\mathrm{avg}}(K)
\;\propto\;
\frac{C_{\mathrm{traj}}+cK}{B_{\mathrm{comp}}}
\left(
\frac{(1-\bar\rho)\sigma^2}{K}
+
\bar\rho\sigma^2
\right).
\label{eq:budgeted-k-variance}
\end{equation}
Treating $K$ as continuous, the minimizer is
\begin{equation}
K^*
=
\sqrt{
\frac{(1-\bar\rho)\,C_{\mathrm{traj}}}
{\bar\rho\,c}
}.
\label{eq:kstar}
\end{equation}
\end{corollary}

\noindent When $\bar\rho$ is large, correlated timesteps provide little additional variance reduction, so the compute-optimal $K$ is small. This is consistent with our empirical finding that $K=4$ is optimal (\Cref{tab:onpolicy_trajactory}).

\subsection{Relationship to single-timestep training (LLaDA)}
\label{app:llada}

In standard masked diffusion training (e.g., LLaDA~\citep{nie2025large}), the state distribution $q(x_t\mid x_0)$ is \emph{fixed} (forward noising of real data) and does not depend on $\theta$. Single-timestep sampling yields an unbiased gradient. In on-policy \emph{distillation}, states come from $q_\theta(x_t)$, which shifts with $\theta$. A single-timestep loss corrects one random step while errors at the other $T-1$ steps compound---the feedback loop described in \Cref{thm:tv-decomposition}. Multi-step trajectory loss controls $K$ terms of the bound simultaneously, breaking the compounding loop.

\section{Implementation Details}
\label{app:implementation}

\subsection{Student architectures}
\Cref{tab:student_architectures} lists the student-model hyperparameters. Each student preserves the corresponding teacher's architecture family and tokenizer while scaling down depth and width to obtain a smaller model. Teachers retain their original architectures .

\begin{table}[h]
\centering
\caption{Hyperparameters of the MDLM and Dream students.}
\label{tab:student_architectures}
\begin{small}
\begin{tabular}{lcc}
\toprule
\textbf{Hyperparameter} & \textbf{MDLM Student} & \textbf{Dream Student} \\
\midrule
Hidden size $d_{\text{model}}$ & 96 & 384 \\
Layers  & 4 & 12 \\
Attention heads & 4 & 4 \\
Sequence length & 1024 & 256 \\
Vocabulary size & 50{,}257 (GPT-2) & 151{,}643 (Qwen) \\
\bottomrule
\end{tabular}
\end{small}
\end{table}

\subsection{Optimization}
\Cref{tab:training_config} lists optimizer-level hyperparameters used for the final student distillation runs. For off-policy training, we follow the corresponding public MDLM and Dream
training-code defaults for optimizer and implementation settings where applicable. The phase-specific learning rates and batch sizes are reported separately in \Cref{tab:distill_run_config}; Dream's off-policy learning rate and batch size are tuned for our student distillation setting.

\begin{table}[h]
\centering
\caption{Optimizer hyperparameters for the final student distillation runs.}
\label{tab:training_config}
\begin{small}
\begin{tabular}{lcc}
\toprule
\textbf{Parameter} & \textbf{MDLM} & \textbf{Dream} \\
\midrule
Optimizer & AdamW & AdamW \\
$\beta_1$ & 0.9 & 0.9 \\
$\beta_2$ & 0.999 & 0.95 \\
$\epsilon$ & $1\times 10^{-8}$ & $1\times 10^{-8}$ \\
Weight decay & 0 & 0.01 \\
Gradient clipping & 1.0 & 1.0 \\
\bottomrule
\end{tabular}
\end{small}
\end{table}

\subsection{Distillation hyperparameters}
\Cref{tab:distill_run_config} summarizes phase-level hyperparameters for the final off-policy and on-policy distillation experiments. For on-policy training, we follow the small-batch, low-learning-rate recipe of \citet{lu2025onpolicydistillation}.

\begin{table}[h]
\centering
\caption{Run-level hyperparameters for the final MDLM and Dream distillation experiments.}
\label{tab:distill_run_config}
\begin{small}
\setlength{\tabcolsep}{5pt}
\begin{tabular}{lcc}
\toprule
\textbf{Hyperparameter} & \textbf{MDLM} & \textbf{Dream} \\
\midrule
Off-policy batch size & 512 & 1024 \\
Off-policy learning rate & $3\times 10^{-4}$ & $1\times 10^{-4}$ \\
On-policy batch size & 16 & 32 \\
On-policy learning rate & $3\times 10^{-8}$ & $3\times 10^{-8}$ \\
Trajectory states $K$ & 4 & 4 \\
Student sampling steps & 1024 & 256 \\
Confidence exponent $\gamma$ & 1.0 & 1.0 \\
Trust-region coefficient $\lambda$ & $3\times 10^{-3}$ & $1\times 10^{-2}$ \\
Trust-region refresh interval & 100 steps & 150 steps \\
\bottomrule
\end{tabular}
\end{small}
\end{table}

\subsection{Compute and wall-clock.}
\Cref{tab:compute} summarizes GPU-hours by phase, reporting MDLM on A100 hardware and Dream on H100 hardware.

\begin{table}[h]
\centering
\caption{Compute cost per phase. MDLM is reported in A100 GPU-hours and Dream in H100 GPU-hours.}
\label{tab:compute}
\begin{small}
\setlength{\tabcolsep}{6pt}
\begin{tabular}{@{}lcc@{}}
\toprule
Phase & \shortstack{MDLM\\(A100)} & \shortstack{Dream\\(H100)} \\
\midrule
Off-policy & $41$ & $1{,}433$ \\
On-policy & $61$ & $394$ \\
Total & $102$ & $1{,}827$ \\
\bottomrule
\end{tabular}
\end{small}
\end{table}

\section{Data Generation and Preprocessing}
\label{app:data}

We use a fully data-free pipeline: every training token comes from teacher generation. For MDLM we sample from the pretrained teacher starting from pure noise with the standard $1024$-step schedule \citep{sahoo2024simple}. For Dream we use unconditional sampling with the same number of steps used at inference.

\subsection{Nested data subsets.}
To rigorously evaluate the scaling behavior of \Cref{sec:challenges}, the MDLM corpora are constructed as strictly nested subsets,
\begin{equation*}
\mathcal{D}_{1\mathrm{M}} \subset \mathcal{D}_{10\mathrm{M}} \subset \mathcal{D}_{30\mathrm{M}} \subset \mathcal{D}_{100\mathrm{M}},
\end{equation*}
so that performance differences across sizes cannot be attributed to distributional shift across regenerations.

\section{Qualitative Samples}
\label{app:samples}

We provide representative generation samples of MDLM to qualitatively compare the baselines against our proposed method.

\subsection{Forward-KL Failure Cases}
The following sample illustrates the failure mode of standard Forward-KL distillation (trained on 10M teacher tokens). The model exhibits severe degeneration, characterized by local repetition loops and semantic incoherence.

\begin{quote}
\small
\textit{``The same thing to us, for us to even more, especially all things out of down. More interesting things that are a FIFA player who loves someone thinks they think you can do it, otherwise we do not build up, obviously. You have to go time, they talk players and qualities we change, play fit around a team, it doesn't go be a part. That will continue to provide the others in relationship so little, and get an opportunity at wow. They don't try to be cautious. If it have someone from him, they need to call you get him into a ball where high to the things they've done with it. You four-year huddle throughout the next day, football and particularly skilled here. These guys got their good game every night and now, and not have a really fair week, but we know we've did the best guys and go around again. At the moment. \"Keep out a simple air camp and climbing because players are going out and everything you have to concentrate on, who kind of the confidence is fine.\"  Although you stand this better to go to the board but maybe would return the game. Everybody knows the players weren't position or be playing with a good group team roster player.  \"The question why to not be the team and my demeanor there's three-game reasons where I think the only team at everyone will be and that is well as a team as an important place, everything has the best attitude on continuity inside the guys are being offered for the players, I've put the expectations and win in the right way yet. They're the reason about it too many things to do the other lot of them and going to show out for what it is great, it or they are still being made there but we plan out to us to stay as with a coaching meeting him and only if the players have been able to get over his career but it's only going to be paid. So we're losing the other team, coming forward out of him,  ''}
\end{quote}

\subsection{Our Method}
The following sample is taken from generation of our method. The model largely avoids the short local repetition loops seen in Forward-KL and maintains a loose film/music theme over a longer span.

\begin{quote}
\small
``
feel that’s not true that we put on a conversation in the first way we’re worried about being all my feeling. But away from what's key thing and other things are supposed to create that space. If true, you just felt smooth it.” Which, he’s not looking at every bigger rewards. That’s a contemporary movies of all what was—and then he leaves me. Or that is reasonable to do my artists to bring him in which I know where these experienced new kids. That’s thought of the DJ song: Hollywood “It’s always hard for the so good think feel great. Still, I’m still having a respect. And that comes when shutting down just getting to me. This seems different piece and sounding bigger or more sensitive things between things. And, a lack of culture as a lot more of the whole chance to use it connected with those right now, the role in the hands of a sequence, which not stupid and instead of art, you can teach artists and some are so deep as a band on recent top project day. It’s excited to be interesting thing with understanding of them constantly and as artistic and develop art and hopefully get one point. People have a real fundamental sense that they tend to learn these advantages, and thoughts where they have the choice for kids? 
 ''
\end{quote}

\section{Broader Impacts}
\label{app:broader_impacts}

This work studies parameter distillation for diffusion language models and does not introduce a new dataset, user-facing application, or deployed system. Its most direct potential positive impact is efficiency. By improving the quality of much smaller DLM students, the method can reduce memory requirements and serving cost relative to using a large teacher model directly. This may make non-autoregressive language generation more practical on constrained hardware, including edge or on-device settings where parallel generation is attractive but large teachers are impractical. The data-free setting may also be useful when the teacher's original training corpus is unavailable or cannot be redistributed.

The same efficiency gains can also have negative impacts. Better small generative models lower the cost of producing large volumes of text, which can facilitate spam, misleading content, impersonation, and other forms of automated misuse. Because our method is explicitly designed to transfer behavior from a teacher model, the student may inherit undesirable properties of the teacher, including toxicity, social biases, unsafe instruction-following behavior, and memorized private information. The data-free setting does not remove these risks: teacher-generated samples and teacher logits can still reflect biases or memorized content from the teacher's original training data. Confidence reweighting may further emphasize high-confidence teacher behavior; if the teacher is confidently wrong or biased, the student can inherit that behavior.

Mitigating these risks requires evaluating distilled students with the same safety, privacy, and fairness checks used for their teachers, rather than assuming that compression or data-free training makes the model safer. Possible mitigations include filtering teacher-generated corpora, measuring toxicity and demographic bias, checking for memorization or privacy leakage, using watermarking or provenance tools for generated text, and applying access controls when the resulting model is capable enough for misuse. Since our experiments focus on modeling quality rather than safety alignment, we view these evaluations as necessary before deploying distilled DLMs in open-ended user-facing settings.

\section{Limitations}
\label{app:limitations}

Our study focuses on data-free parameter distillation for masked diffusion language models. While we evaluate two representative regimes, MDLM-110M to a 10.5M student and Dream-7B to a 93M student, both students preserve the architecture family and tokenizer of their corresponding teachers. We therefore do not claim that the same recipe directly transfers to other discrete diffusion parameterizations, architecture-mismatched students, tokenizer-mismatched distillation, or continuous diffusion models.

The method also introduces additional computation and hyperparameters beyond standard off-policy distillation. On-policy training requires student rollouts and teacher evaluations at multiple trajectory states, making the teacher-query cost non-negligible, especially for large teachers such as Dream-7B. The final recipe depends on choices such as the number of trajectory states $K$, confidence exponent $\gamma$, trust-region coefficient $\lambda$, refresh interval $\Delta$, and divergence family. We ablate several of these choices, but do not exhaustively explore adaptive schedules, alternative trust-region divergences, or automatic phase-switching criteria. 

We focus on masked DLMs because they currently represent the more scalable and practically relevant DLM paradigm. At comparable model scales, masked DLMs generally outperform uniform DLMs, as illustrated by MDLM and Duo, while recent efforts to scale DLMs to frontier sizes, such as the 100B-parameter LLaDA 2, have adopted masked diffusion. Moreover, uniform DLMs such as Duo typically perform diffusion in continuous representation spaces, making existing distillation methods for continuous diffusion models more directly applicable to them. In contrast, masked DLMs operate over discrete masked-token states, introducing a distinct and comparatively underexplored distillation problem. We therefore chose masked DLMs as the primary scope of this study, while acknowledging that extending our analysis and framework to other DLM variants remains important future work.

\newpage
\section*{NeurIPS Paper Checklist}
\begin{enumerate}

\item {\bf Claims}
    \item[] Question: Do the main claims made in the abstract and introduction accurately reflect the paper's contributions and scope?
    \item[] Answer: \answerYes{} % Replace by \answerYes{}, \answerNo{}, or \answerNA{}.
    \item[] Justification: The main claims made in the abstract and introduction accurately reflect the paper’s contributions and scope.

\item {\bf Limitations}
    \item[] Question: Does the paper discuss the limitations of the work performed by the authors?
    \item[] Answer: \answerYes{} % Replace by \answerYes{}, \answerNo{}, or \answerNA{}.
    \item[] Justification: We include a through limitation discussion in \Cref{app:limitations} 

\item {\bf Theory assumptions and proofs}
    \item[] Question: For each theoretical result, does the paper provide the full set of assumptions and a complete (and correct) proof?
    \item[] Answer: \answerYes{} % Replace by \answerYes{}, \answerNo{}, or \answerNA{}.
    \item[] Justification: For each theoretical result, the paper provides the full set of assumptions and a complete (and correct) proof.

    \item {\bf Experimental result reproducibility}
    \item[] Question: Does the paper fully disclose all the information needed to reproduce the main experimental results of the paper to the extent that it affects the main claims and/or conclusions of the paper (regardless of whether the code and data are provided or not)?
    \item[] Answer: \answerYes{} % Replace by \answerYes{}, \answerNo{}, or \answerNA{}.
    \item[] Justification: The paper fully disclose all the information needed to reproduce the main experimental results of the paper, and we will release all our code to assist the reproducibility of our experimental results.

\item {\bf Open access to data and code}
    \item[] Question: Does the paper provide open access to the data and code, with sufficient instructions to faithfully reproduce the main experimental results, as described in supplemental material?
    \item[] Answer: \answerYes{} % Replace by \answerYes{}, \answerNo{}, or \answerNA{}.
    \item[] Justification: We will release our code, models and data.

\item {\bf Experimental setting/details}
    \item[] Question: Does the paper specify all the training and test details (e.g., data splits, hyperparameters, how they were chosen, type of optimizer) necessary to understand the results?
    \item[] Answer: \answerYes{} % Replace by \answerYes{}, \answerNo{}, or \answerNA{}.
    \item[] Justification: We specify all the training and test details in the paper.

\item {\bf Experiment statistical significance}
    \item[] Question: Does the paper report error bars suitably and correctly defined or other appropriate information about the statistical significance of the experiments?
    \item[] Answer: \answerNo{} % Replace by \answerYes{}, \answerNo{}, or \answerNA{}.
    \item[] Justification: DLM distillation with large teachers consumes substantial computational resources, making it impractical to conduct multiple full training runs to obtain error bars. 

\item {\bf Experiments compute resources}
    \item[] Question: For each experiment, does the paper provide sufficient information on the computer resources (type of compute workers, memory, time of execution) needed to reproduce the experiments?
    \item[] Answer: \answerYes{} % Replace by \answerYes{}, \answerNo{}, or \answerNA{}.
    \item[] Justification: We provided sufficient information on the computer resources in the main text and appendix.
    
\item {\bf Code of ethics}
    \item[] Question: Does the research conducted in the paper conform, in every respect, with the NeurIPS Code of Ethics \url{https://neurips.cc/public/EthicsGuidelines}?
    \item[] Answer: \answerYes{} % Replace by \answerYes{}, \answerNo{}, or \answerNA{}.
    \item[] Justification: The research conducted in the paper conformed, in every respect, with the NeurIPS Code of Ethics.

\item {\bf Broader impacts}
    \item[] Question: Does the paper discuss both potential positive societal impacts and negative societal impacts of the work performed?
    \item[] Answer: \answerYes{} % Replace by \answerYes{}, \answerNo{}, or \answerNA{}.
    \item[] Justification: We discuss potential societal impacts in the appendix.
    
\item {\bf Safeguards}
    \item[] Question: Does the paper describe safeguards that have been put in place for responsible release of data or models that have a high risk for misuse (e.g., pre-trained language models, image generators, or scraped datasets)?
    \item[] Answer: \answerYes{} % Replace by \answerYes{}, \answerNo{}, or \answerNA{}.
    \item[] Justification: We describe safeguards for responsible release of models in appendix.

\item {\bf Licenses for existing assets}
    \item[] Question: Are the creators or original owners of assets (e.g., code, data, models), used in the paper, properly credited and are the license and terms of use explicitly mentioned and properly respected?
    \item[] Answer: \answerYes{} % Replace by \answerYes{}, \answerNo{}, or \answerNA{}.
    \item[] Justification: We properly credited the creators or original owners of assets (e.g., code, data, models), used in the paper and conformed the license and terms.

\item {\bf New assets}
    \item[] Question: Are new assets introduced in the paper well documented and is the documentation provided alongside the assets?
    \item[] Answer: \answerYes{} % Replace by \answerYes{}, \answerNo{}, or \answerNA{}.
    \item[] Justification: We document all assets.

\item {\bf Crowdsourcing and research with human subjects}
    \item[] Question: For crowdsourcing experiments and research with human subjects, does the paper include the full text of instructions given to participants and screenshots, if applicable, as well as details about compensation (if any)? 
    \item[] Answer: \answerNA{} % Replace by \answerYes{}, \answerNo{}, or \answerNA{}.
    \item[] Justification: The paper does not involve crowdsourcing nor research with human subjects.

\item {\bf Institutional review board (IRB) approvals or equivalent for research with human subjects}
    \item[] Question: Does the paper describe potential risks incurred by study participants, whether such risks were disclosed to the subjects, and whether Institutional Review Board (IRB) approvals (or an equivalent approval/review based on the requirements of your country or institution) were obtained?
    \item[] Answer: \answerNA{} % Replace by \answerYes{}, \answerNo{}, or \answerNA{}.
    \item[] Justification:  The paper does not involve crowdsourcing nor research with human subjects.

\item {\bf Declaration of LLM usage}
    \item[] Question: Does the paper describe the usage of LLMs if it is an important, original, or non-standard component of the core methods in this research? Note that if the LLM is used only for writing, editing, or formatting purposes and does \emph{not} impact the core methodology, scientific rigor, or originality of the research, declaration is not required.
    %this research? 
    \item[] Answer: \answerNA{} % Replace by \answerYes{}, \answerNo{}, or \answerNA{}.
    \item[] Justification: The core method development in this research does not involve LLMs as any important, original, or non-standard components.

\end{enumerate}

\end{document}